\documentclass[onefignum,onetabnum]{siamart220329}

\usepackage{enumitem}
\usepackage{algorithm}
\usepackage{algpseudocode}
\usepackage{amsmath}
\usepackage{multirow}

\usepackage{amsfonts}
\usepackage{graphics}
\usepackage{xcolor}
\usepackage{amssymb}
\usepackage{psfrag}
\usepackage{graphicx}
\usepackage[utf8]{inputenc}
\usepackage{bm}
\usepackage{mathtools}
\usepackage{epstopdf}
\usepackage{setspace}
\usepackage{amsmath,amssymb,bbm}
\usepackage[utf8]{inputenc}
\usepackage{url}
\usepackage{color}
\usepackage{booktabs}

\newsiamremark{remark}{Remark}
\newsiamremark{hypothesis}{Hypothesis}
\crefname{hypothesis}{Hypothesis}{Hypotheses}
\newsiamthm{claim}{Claim}
\newsiamthm{assum}{Assumption}
\newsiamthm{scheme}{Scheme}
\newsiamthm{lem}{Lemma}
\newsiamthm{thm}{Theorem}
\headers{A Conditional Diffusion Model for Learning SDEs}{Y.~Liu, Y.~Chen, D.~Xiu, G.~Zhang}

\title{A Training-Free Conditional Diffusion Model for Learning Stochastic Dynamical Systems\thanks{The work of Y. Liu and G. Zhang was partially supported by the U.S. Department of Energy, Office of Science, Office of Advanced Scientific Computing Research, Applied Mathematics program, under the contracts ERKJ388 and ERKJ443. 
ORNL is operated by UT-Battelle, LLC., for the U.S. Department of Energy under Contract DE-AC05-00OR22725. The work of Y. Chen and D.Xiu was partially supported by AFOSR FA9550-22-1-0011.}}

\author{
Yanfang Liu\thanks{Department of Mathematics, Middle Tennessee State University, Murfreesboro, TN 37132 (yanfang.liu@mtsu.edu).}
\and 
Yuan Chen\thanks{Department of Mathematics, Ohio State University, Columbus, OH 43210 ({chen.11050@osu.edu}, xiu.16@osu.edu).}
 \and 
Dongbin Xiu\footnotemark[3]
\and
Guannan Zhang\thanks{Computer Science and Mathematics Division, Oak Ridge National Laboratory, Oak Ridge, TN 37831 (zhangg@ornl.gov).}
}

\usepackage{bbm}

\allowdisplaybreaks
\begin{document}

\maketitle

\begin{abstract}
This study introduces a training-free conditional diffusion model for learning unknown stochastic differential equations (SDEs) using data. The proposed approach addresses key challenges in computational efficiency and accuracy for modeling SDEs by utilizing a score-based diffusion model to approximate their stochastic flow map. Unlike the existing methods, this technique is based on an analytically derived closed-form exact score function, which can be efficiently estimated by Monte Carlo method using the trajectory data, and eliminates the need for neural network training to learn the score function. By generating labeled data through solving the corresponding reverse ordinary differential equation, the approach enables supervised learning of the flow map. Extensive numerical experiments across various SDE types, including linear, nonlinear, and multi-dimensional systems, demonstrate the versatility and effectiveness of the method. The learned models exhibit significant improvements in predicting both short-term and long-term behaviors of unknown stochastic systems, often surpassing baseline methods like GANs in estimating drift and diffusion coefficients.
\end{abstract}



\section{Introduction}\label{sec:intro}
Stochastic differential equations (SDEs) play a pivotal role in scientific and engineering simulations, particularly when modeling large ensembles of particles. These equations find diverse applications across fields such as plasma physics, where they describe electron and ion interactions with electromagnetic fields, and fluid mechanics, where they model the dispersion of pollutants in oceans and the atmosphere. SDEs also prove valuable in simulating biological and chemical systems. From a mathematical standpoint, SDEs offer a discrete particle representation that corresponds to the continuous Fokker-Planck equation, underpinning Monte Carlo methods for solving partial differential equations. 
While highly valuable, the use of SDEs faces a significant challenge: for many complex systems it is difficult, if not impossible, to derive accurate SDE models that capture all the important physics embedded in the systems.

Recent years have seen a growing interest in developing data-driven approaches for uncovering unknown dynamical systems. The primary objective is to unearth the core principles or mathematical formulations underlying observational data, enabling the creation of effective predictive models for unexplored dynamics. When dealing with stochastic systems, the presence of data noise and the inability to directly observe the system's inherent randomness present considerable obstacles in understanding the system. Most existing approaches concentrate on analyzing Itô-type SDEs. These methods utilize various techniques, including Gaussian processes \cite{pmlr-v1-archambeau07a,DARCY2023133583,https://doi.org/10.1002/andp.201800233}, polynomial approximations \cite{LI2021132830,benjaminfom}, and deep neural networks \cite{CHEN2023133559,doi:10.1137/20M1360153,doi:10.1137/21M1413018}. A recent development involves the extension of the deterministic flow map approach to stochastic scenarios, incorporating generative models like generative adversarial networks (GANs) \cite{CHEN2024112984,QIN2019620}, autoencoders \cite{xu2023,CHENG2024116793}, and normalizing flows \cite{doi:10.1137/23M1585635}.

Despite the success of generative models
in learning SDEs, there is a significant challenge in training the generative models to achieve the desired accuracy. Due to the lack of labeled data, training generative models is usually classified as unsupervised learning. 
Various unsupervised loss functions have been defined to train generative models, including adversarial loss for GANs \cite{NIPS2014_5ca3e9b1}, the maximum likelihood loss for normalizing flows \cite{kobyzev2020normalizing}, and the score matching losses for diffusion models 
\cite{JMLR:v6:hyvarinen05a,10.1162/NECO_a_00142,pmlr-v115-song20a}. There are several computational issues resulted from the unsupervised training nature of generative models. For example, the training of GANs may suffer from mode collapse, vanishing gradients, and training instability \cite{DBLP:conf/nips/SalimansGZCRCC16}. The maximum likelihood loss used in normalizing flows requires efficient computation of the determinant of the Jacobian matrix, which places restrictions on networks' architectures \cite{Kobyzev_2021}.

In this work, we propose a training-free conditional diffusion model that enables supervised learning of unknown SDEs. Diffusion models have been successfully used in a variety of applications, including image processing \cite{nichol2021improved,song2021score,song2019generative,rombach2022high,dhariwal2021diffusion,ho2020denoising}, natural language processing \cite{li2022diffusion,yu2022latent,chen2022analog,gong2022diffuseq}, and uncertainty quantification \cite{liu2024diffusion,bao2024score,DBUQ,li2023diffusion,score_uq}. Unlike existing diffusion models \cite{song2021score,nichol2021improved,yang2022diffusion}, our method does not need to train neural networks to learn the conditional score function. Specifically, we derived the closed form of the exact score function for the conditional diffusion model used to approximate the stochastic flow map of the target SDE. In fact, the score function can be represented as an expectation with respect to the conditional distribution defined by the flow map. However, since we use trajectories of the target SDE as observation data, we do not have a large number of samples from each conditional distribution. Instead, the trajectory data can be treated as samples from the joint distribution of the input and the output of the flow map. Thus, we propose an approximate score function that involves a weighted expectation with respect to the joint distribution, and use Monte Carlo estimators (based on the trajectory data) to approximate the expectation. This approach allows us to directly generate samples of the target flow map. With the estimated score function, we then generate labeled data by solving the corresponding reverse ordinary differential equation, instead of the reverse SDE for the diffusion model. Following this, we employ the generated labeled data to train a simple fully connected neural network to learn the flow map via supervised learning. Our method shows remarkable accuracy in predicting both short-term and long-term behaviors of stochastic systems, often outperforming baseline methods such as GANs \cite{CHEN2024112984} in terms of drift and diffusion coefficient estimation, as well as in predicting mean and standard deviation at termination times.

The rest of this paper is organized as follows. In Section \ref{sec:problem}, we briefly introduce the learning of unknown stochastic dynamical systems under consideration. In Section \ref{sec:method}, we provide a comprehensive discussion of a conditional score-based generative diffusion model to learn the flow map. Finally in  Section \ref{sec:results}, we demonstrate the performance of our method by applying it to a set of analytical stochastic systems and real applications.

%
\section{Problem setting}\label{sec:problem}
Let $\{\Omega,\mathcal{F},\mathbb{P},\{\mathcal{F}_t\}_{0\leq t\leq
T}\}$ be a complete, filtered probability space on which a standard
$m$-dimensional Brownian motion $  W_t$ is defined, such that
$\{\mathcal{F}_t\}_{0\leq t\leq T}$ is the natural filtration of the
Brownian motion $  W_t$ and all the $\mathbb{P}$-null sets are augmented to each
$\sigma$-field $\mathcal{F}_t$. 
In the probability space $\{\Omega,\mathcal{F},\mathbb{P},\{\mathcal{F}_t\}_{0\leq t\leq
T}\}$, we define the following $d$-dimensional stochastic differential equation:
\begin{equation}\label{eq: problem_SDE}
     d{ {X}}_{t}= { {a}}({ {X}}_{t}) dt + { {b}}({ {X}}_{t})\, d{{  {W}}}_{t},
\end{equation}
where $a:\mathbb{R}^{d}\rightarrow \mathbb{R}^{d}$ is referred to as the drift coefficient, $b:\mathbb{R}^{d}\rightarrow \mathbb{R}^{d \times m}$ with $m\leq d$ is referred to as the diffusion coefficient.
Assuming $a(x)$ and $b(x)$ satisfy the Lipschitz condition and the linear growth condition, the SDE in Eq.~\eqref{eq: problem_SDE} has a unique solution 
\begin{equation}\label{eq: SDE_sln}
    { {X}}_{t+\Delta t}^{t,x} = { {x}} + \int_{t}^{t+\Delta t}   {{a}}({ {X}}_{s}^{t,x}) ds + \int_{t}^{t+\Delta t}   {{  b}}({ {X}}_{s}^{t,x})  d{  {W}}_{s} \; \text{ with } \;
    X_t = x,
\end{equation}
for any $0 \le t \le T$ and $\Delta t > 0$, where ${ {X}}_{t+\Delta t}^{t,x}$ is the solution of the SDE at $t+\Delta t$ conditional on $X_t = x$ and
$\int_{t}^{t+\Delta t}b(X_{s}^{t,x})dW_{s}$ is an It\^{o} integral. 
Because both the drift and diffusion terms do not explicitly depend on the time $t$, the probability distribution of increment $X_{t+\Delta t}^{t,x} - x$ only depends on the starting location $x$ and the time step $\Delta t$, regardless of the time instant $t$ in Eq.~\eqref{eq: SDE_sln}. 
Therefore, the representation in Eq.~\eqref{eq: SDE_sln} defines a 
stochastic flow map:
\begin{equation}\label{eq:truefm}
 F_{\Delta t}(x, \omega) := {X}^{x}_{\Delta t} - { x },
\end{equation}
where we omit the dependence of ${X}^{x}_{\Delta t}$ on $t$,
$\omega$ represents the sample from the probability space $\{\Omega,\mathcal{F},\mathbb{P},\{\mathcal{F}_t\}_{0\leq t\leq
T}\}$, and $F_{\Delta t}(x, \omega)$ is independent of $t \in [0,T]$. 

{ The objective} of this work is to build a conditional generative model, denoted by
$G_\theta({x},z)$, to approximate the exact flow map $F_{\Delta t}( x, \omega)$ in Eq.~\eqref{eq:truefm}, i.e., 
\begin{equation}\label{eq:gf}
    G_\theta(x,z) \approx F_{\Delta t}( x,\omega), 
\end{equation}
where $x$ denotes a sample of $X_{t}$ for $t \in [0,T]$, $z$ denotes a sample from the standard normal distribution, and $\theta$ denotes the set of trainable parameters of the conditional generative model. For notational simplicity, we omit the subscript $\Delta t$ in $G_\theta(x,z)$ in the rest of this paper. Also, we introduce a uniform temporal mesh
\begin{equation}\label{eq:time}
    \mathcal{T} := \{t_n : t_n = n \Delta t \; \text{ for }\; n = 0, 1, \ldots, N_T\},
\end{equation}
where $\Delta t = T/N_T$, and we focus on approximating the flow map $F_{\Delta t}(x, \omega) $ defined by the time step $\Delta t$ associated with the temporal mesh $\mathcal{T}$.
Once the generative model $G_\theta(x, z)$ is well trained, it can be used to efficiently generate unlimited trajectories of the SDE in Eq.~\eqref{eq: problem_SDE} in an auto-regressive manner.

The observation data set of the target SDE includes $H\geq 1$ solution trajectories of $ { {X}}_{t}$ at discrete time instants on the uniform temporal mesh $\mathcal{T}$ in Eq.~\eqref{eq:time}, denoted by
\begin{equation} \label{eq:raw_trajectory}
     X_{t_0}^{(i)}, X_{t_1}^{(i)}, \cdots, 
     X_{t_L}^{(i)}, \qquad i=1,\cdots, H,
\end{equation}
where $\{t_0, t_1, \ldots, t_L\} \in \mathcal{T}$, $L+1$ denotes the length of the $i$-th trajectory, and the initial state $X_{t_0}^{(i)}$ is a sample from a given initial distribution $p_{X_0}(x)$. 
The trajectory data in 
Eq.~\eqref{eq:raw_trajectory} can be separated and regrouped to obtain data pairs that can be used to describe the input-output relationship of the flow map $F_{\Delta t}(x,\omega)$, i.e.,
\begin{equation}\label{eq:pairs}
    x_m := X_{t_l}^{(i)}\, \text{ and }\,
    \Delta x_m := X_{t_{l+1}}^{(i)} - X_{t_{l}}^{(i)}, 
\end{equation}
for $m = l\times H+i$ with $l = 0, \ldots, L-1, i = 1, \ldots, H$, which leads to a total of $M=HL$ adjacent data pairs. We denote the collection of the paired samples as the
observation data set for the flow map, i.e., 
\begin{equation}\label{eq:obs}
    \mathcal{D}_{\rm obs} := 
    \left\{ (x_m, \Delta x_m) \, | \, m=1,\cdots,M = HL \right\}.
\end{equation}
The next section will describe how to use the observation data set $\mathcal{D}_{\rm obs}$ in Eq.~\eqref{eq:obs} to perform a supervised training of the desired generative model in Eq.~\eqref{eq:truefm}

\section{The conditional diffusion model for supervised learning of the flow map}\label{sec:method}
We now describe the details of the proposed training-free conditional diffusion model for learning the flow map of interest. Section \ref{sec:diff_over} introduces the conditional score-based diffusion model; Section \ref{sec:mc} introduces the definition of the conditional score function as well as its training-free approximation scheme; Section \ref{sec:learning} describes how to use the proposed conditional diffusion model to generate labeled data for supervised training of the desired conditional generative model $G_\theta(\cdot, \cdot)$ in Eq.~\eqref{eq:gf}.

\subsection{The conditional score-based diffusion model}\label{sec:diff_over}
We intend to define a score-based diffusion model to represent a transport map from a standard normal random variable, denoted by $Z \sim \mathcal{N}(0, \mathbf{I}_d)$, and the random variable ${X}^{x}_{\Delta t} - { x }$ in Eq.~\eqref{eq:truefm} conditional on $X_t = x$ for any $t \in [0,T]$. It is referred to as a conditional diffusion model because the generated probability distribution also depends on the conditional state $x$. 
 The diffusion model includes a forward SDE and a reverse SDE, both of which are defined in a bounded temporal domain $\tau \in [0,1]$. The forward SDE is defined by
\begin{equation}\label{eq:forward}
 dZ_\tau^x = b(\tau) Z_{\tau}^x d\tau + \sigma(\tau) dW_\tau\; \text{ with }\; Z_{0}^x= {X}^{x}_{\Delta t} - x,
\end{equation}
where $Z_0^x$ is the target random variable ${X}^{x}_{\Delta t} - x$ for a fixed $x$. Note that the entire forward SDE for the diffusion model is conditional on $X_t = x$ for any $t \in [0,T]$. When properly choosing the drift and diffusion coefficients, the forward SDE can transport the initial distribution $p_{Z_0}(z_0)$ to the standard normal distribution at $\tau=1$. In this work, we use the following definitions of $b(\tau)$ and $\sigma(\tau)$ in Eq.~(\ref{eq:forward}):
\begin{equation}\label{eq:cof}
\begin{aligned}
b(\tau) = \frac{{\rm d} \log \alpha_\tau}{{\rm d} \tau} \;\;\; \text{ and }\;\;\; \sigma^2(\tau) = \frac{{\rm d} \beta_\tau^2}{{\rm d}\tau} - 2 \frac{{\rm d}\log \alpha_\tau}{{\rm d}\tau} \beta_\tau^2,
\end{aligned}
\end{equation}
where $\alpha_\tau$ and $\beta_\tau$ are set as
\begin{equation}\label{eq:ab}
\alpha_\tau = 1-\tau, \;\; \beta^2_\tau = \tau \;\; \text{ for } \;\; \tau \in [0,1).
\end{equation}
Because the forward process in Eq.~\eqref{eq:forward} is a \textit{linear} SDE, its solution is 
\begin{equation}\label{SDE_solu}
    Z_{\tau}^x = Z_{0}^x\exp\left[\int_0^\tau b(s)ds\right] + \int_0^\tau\exp\left[\int_s^\tau b(r) dr\right]\sigma(s)dW_s.
\end{equation}
Substituting $b(\tau)$ and $\sigma(\tau)$ defined in Eq.~\eqref{eq:cof} into Eq.~\eqref{SDE_solu}, we have the conditional PDF $q_{Z_{\tau}^x|Z_{0}^x}(z_{\tau}^x|z_{0}^x)$ for a fixed $z_{0}^x$ is a Gaussian distribution: 
\begin{equation}\label{eq:gauss}
q_{Z_{\tau}^x|Z_{0}^x}(z_{\tau}^x|z_{0}^x) = \psi_{(\alpha_\tau z_{0}^x, \beta_\tau^2 \mathbf{I}_d)}(z_{\tau}^x|z_{0}^x), 
\end{equation}
where $\psi_{(\alpha_\tau z_{0}^x, \beta_\tau^2 \mathbf{I}_d)}(\cdot)$ is the standard normal PDF with mean $\alpha_\tau z_{0}^x$ and covariance matrix $\beta_\tau^2 \mathbf{I}_d$. 
Then the asymptotic limit of $q_{Z_{\tau}^x|Z_{0}^x}(z_{\tau}^x|z_{0}^x)$ as $\tau \rightarrow 1$ is the standard normal distribution, which means forward SDE can transport any initial distribution to the standard normal distribution within a bounded pseudo-temporal domain $[0,1)$.

The corresponding reverse SDE, that is used to generate new samples of the target distribution, is defined by
\begin{equation}\label{eq:backward}
d{Z}_\tau^x = \left[ b(\tau){Z}_\tau^x - \sigma^2(\tau) S(Z_\tau^x, \tau)\right] d\tau + \sigma(\tau) d{B}_\tau \quad \text{with}\; Z_1^x =  Z \sim \mathcal{N}(0, \mathbf{I}_d),
\end{equation}
where ${B}_\tau$ is the reverse-time Brownian motion and $S(z_\tau^x, \tau)$ is the score function defined by
\begin{equation}\label{eq:exact_score}
S(z_{\tau}^x, \tau) := \nabla_z \log p_{Z_\tau^x}({z}_\tau^x),
\end{equation}
where $p_{Z_\tau^x}({z}_\tau^x)$ is the PDF of the state $Z_\tau^x$ in the forward SDE in Eq.~\eqref{eq:forward}.
The reverse SDE in Eq.~(\ref{eq:backward}) performs as a denoiser, which can transform the standard normal distribution of $p_{Z_1^x}(z_1^x)$ to the target distribution $p_{Z_0^x}(z_0^x)$ for $Z_0^x = {X}^{x}_{\Delta t} - x$. If the score function is known, then we can solve the reverse SDE in Eq.~(\ref{eq:backward}) to generate unlimited amount of samples of the flow map in Eq.~\eqref{eq:truefm}. 
The standard score-based diffusion model uses a neural network to learn the score function, which is computationally intensive because it requires to solve the reverse SDE to generate each sample of the target distribution. Moreover, when estimating the score function, the neural network is trained in an unsupervised manner due to the lack of labeled data, which requires storing a large number of trajectories of the forward SDE to be used to train the score function approximation.

\subsection{Training-free score estimation}\label{sec:mc}
We introduce the score estimation approach in this subsection. To proceed, we first write 
out the PDF $p_{Z_\tau^x}({z}_\tau^x)$ in Eq.~\eqref{eq:exact_score} as the following integral form
\begin{equation}\label{eq:pdf}
    q_{Z_\tau^x}(z_\tau^x)= \int_{\mathbb{R}^d} q_{Z_\tau^x, Z_0^x}(z_\tau^x, z_0^x) dz_0^x = \int_{\mathbb{R}^d} q_{{Z}_\tau^x | Z_0^x}(z_\tau^x|z_0^x) q_{Z_0^x}(z_0^x) dz_0^x,
\end{equation}
where $q_{{Z}_\tau^x | Z_0^x}(z_\tau^x|z_0^x)$ is defined in Eq.~\eqref{eq:gauss}, and 
$q_{Z_0^x}(z_0^x)$ is represented by
\begin{equation}\label{eq:q0}
    q_{Z_0^x}(z_0^x) = p_{X^{x}_{\Delta t}-x}(z_0^x),
\end{equation}
according to the definition $Z_{0}^x= {X}^{x}_{\Delta t} - x$ given in Eq.~\eqref{eq:forward}. Substituting Eq.~\eqref{eq:pdf} and Eq.~\eqref{eq:q0} into
Eq.~\eqref{eq:exact_score}, we have
\begin{equation}\label{eq:score11}
\begin{aligned}
 S(z_{\tau}^x, \tau)
  =\, & \nabla_z \log \left(\int_{\mathbb{R}^d} q_{{Z}_{\tau}^x | Z_{0}^x}({z}_{\tau}^x | z_{0}^x) q_{Z_{0}^x}(z_{0}^x) dz_{0}^x\right)\\
 =\, & \frac{1}{\int_{\mathbb{R}^d} q_{{Z}_{\tau}^x | Z_{0}^x}({z}_{\tau}^x | \bar{z}^x_{0}) q_{Z_{0}^x}(
\bar{z}^x_{0}) d\bar{z}^x_{0}}   \int_{\mathbb{R}^d}  - \frac{z_{\tau}^x - \alpha_\tau z_{0}^x}{\beta^2_\tau}q_{Z_{\tau}^x|Z_{0}^x}({z}_{\tau}^x | z_{0}^x) q_{Z_{0}^x}(
z_{0}^x) dz_{0}^x\\[5pt]
=\, & \int_{\mathbb{R}^d}  - \frac{z_{\tau}^x- \alpha_\tau z_{0}^x}{\beta^2_\tau} w({z}_{\tau}^x,  z_{0}^x)  dz_{0}^x,\\
\end{aligned}
\end{equation}
where the weight function $w({z}_{\tau}^x,  z_{0}^x)$ is defined by
\begin{equation}\label{eq:weight}
w({z}_{\tau}^x,  z_{0}^x) := \frac{ q_{Z_{\tau}^x|Z_{0}^x}({z}_{\tau}^x | z_{0}^x) }{\displaystyle \int_{\mathbb{R}^d} q_{Z_{\tau}^x|Z_{0}^x}({z}_{\tau}^x | \bar{z}^x_{0})  q_{Z_{0}^x}(
\bar{z}^x_{0}) d \bar{z}^x_{0}},
\end{equation}
satisfying that $\int_{\mathbb{R}^d}w({z}_{\tau}^x,  z_{0}^x) q_{Z_{0}^x}(z_{0}^x) dz_{0}^x = 1$. 

Instead of training a neural network to learn the score function, we use Monte Carlo estimation to directly approximate the integrals in Eq.~\eqref{eq:score11}. Unlike unconditional diffusion models where we have samples from $q_{Z_0^x}(z_0^x)$ as the observation data, we do {\em not} have a large number of samples from $Z_{0}^x= {X}^{x}_{\Delta t} - x$ for any fixed $x$ due to the way the observation data set $\mathcal{D}_{\rm obs}$ in Eq.~\eqref{eq:obs} is constructed. On the other hand, there may be many samples in $\mathcal{D}_{\rm obs}$ are located within a small neighborhood of $x$. Hence, we need to develop an approximation to the initial distribution $q_{Z_0^x}(z_0^x)$. Specifically, we can rewrite $q_{Z_{0}^x}(z_{0}^x)$ as 
\begin{equation}\label{eq:q0delta}
    q_{Z_{0}^x}(z_{0}^x) = \int_{\mathbb{R}^d} q_{Z_{0}^{\tilde x}}(z_{0}^{\tilde{x}}) \delta(\tilde x - x) d\tilde x,
\end{equation}
where $\delta(\cdot)$ is the Dirac delta function. If we treat $\delta(\tilde x - x)$ as a posterior distribution of $\tilde x$ conditional on $x$, then we can define an approximation scheme of $\delta(\tilde x - x)$ using the Bayesian formula,
\begin{equation}\label{eq:bayes}
    \delta(\tilde x - x) \approx \psi_{(x,\,\nu^2\mathbf{I}_d)}(\tilde{x})\, p_X(\tilde x),
\end{equation}
where $\psi_{(x,\,\nu^2\mathbf{I}_d)}(\tilde{x})$ is the Gaussian likelihood with mean $x$ and standard deviation $\nu >0$, and $p_X(\tilde x)$ is the prior distribution. 
It is easy to see that the posterior distribution $\psi_{(x,\,\nu^2\mathbf{I}_d)}(\tilde{x})\, p_X(\tilde x)$ converges to $\delta(\tilde x - x)$ as $\nu \rightarrow 0$ in Eq.~\eqref{eq:bayes}. Substituting Eq.~\eqref{eq:bayes} into Eq.~\eqref{eq:q0delta}, we have an approximation of $q_{Z_{0}^x}(z_{0}^x)$, i.e.,
\begin{equation}\label{eq:q0app}
    q_{Z_{0}^x}(z_{0}^x) \approx \widehat{q}_{Z_{0}^x}(z_{0}^x) := 
    \int_{\mathbb{R}^d} \psi_{(x,\,\nu^2\mathbf{I}_d)}(\tilde{x})\, q_{Z_{0}^{\tilde x}}(z_{0}^{\tilde{x}})\,  p_X(\tilde x)  d\tilde x,
\end{equation}
where the product $q_{Z_{0}^{\tilde x}}(z_{0}^{\tilde{x}})\,  p_X(\tilde x)$ is treated as a joint PDF of $(\tilde x, z_0^{\tilde x})$, denoted by
\begin{equation}\label{eq:jointpdf}
    q_{X,Z_0}(\tilde{x},z_0^{\tilde{x}}) := q_{Z_{0}^{\tilde x}}(z_{0}^{\tilde{x}})\,  p_X(\tilde x).
\end{equation}
Substituting Eq.~\eqref{eq:q0app} into Eq.~\eqref{eq:score11}, we have an approximate score function as follows:
\begin{equation}\label{eq:appscore}
\begin{aligned}
 \widehat{S}(z_{\tau}^x, \tau)
  :=\, & \nabla_z \log \left(\int_{\mathbb{R}^d}\int_{\mathbb{R}^d} \left[q_{{Z}_{\tau}^{\tilde x} | Z_{0}^{\tilde x}}({z}_{\tau}^{x} | z_{0}^{\tilde x})\, \psi_{(x,\,\nu^2\mathbf{I}_d)}(\tilde{x})\right]\, q_{X,Z_0}(\tilde{x},z_0^{\tilde{x}})\, dz_{0}^{\tilde x}d\tilde{x}\right) \\[5pt]
 =\, & \dfrac{\displaystyle \int_{\mathbb{R}^d} \int_{\mathbb{R}^d}  - \frac{z_{\tau}^{x} - \alpha_\tau z_{0}^{\tilde x}}{\beta^2_\tau} \, \left[q_{Z_{\tau}^{\tilde x}|Z_{0}^{\tilde x}}({z}_{\tau}^{x} | z_{0}^{\tilde x}) \, \psi_{(x,\,\nu^2\mathbf{I}_d)}(\tilde{x}) \right] q_{X,Z_0}(\tilde{x},z_0^{\tilde{x}})\, dz_{0}^{\tilde x}d\tilde{x}}{\displaystyle \int_{\mathbb{R}^d} \int_{\mathbb{R}^d} \left[q_{Z_{\tau}^{\bar x}|Z_{0}^{\bar x}}({z}_{\tau}^{x} | \bar{z}_{0}^{\bar x}) \, \psi_{(x,\,\nu^2\mathbf{I}_d)}(\bar{x}) \right] q_{X,Z_0}(\bar{x},\bar{z}_0^{\bar{x}})\, d\bar{z}_{0}^{\bar x}d\bar{x}}\\[5pt]
=\, & \int_{\mathbb{R}^d}\int_{\mathbb{R}^d}  - \frac{z_{\tau}^{x}- \alpha_\tau z_{0}^{\tilde x}}{\beta^2_\tau} \widehat{w}({z}_{\tau}^{x},  z_{0}^{\tilde x})  \,q_{X,Z_0}(\tilde{x},{z}_0^{\tilde{x}}) \,dz_{0}^{\tilde x}\, d{\tilde x},\\
\end{aligned}
\end{equation}
where the weight function $\widehat{w}({z}_{\tau}^{\tilde x},  z_{0}^{\tilde x})$ is defined by
\begin{equation}\label{eq:app_weight}
  \widehat{w}({z}_{\tau}^{x},  z_{0}^{\tilde x})  := \frac{ q_{Z_{\tau}^{\tilde x}|Z_{0}^{\tilde x}}({z}_{\tau}^{x} | z_{0}^{\tilde x}) \, \psi_{(x,\,\nu^2\mathbf{I}_d)}(\tilde{x})}{\displaystyle \int_{\mathbb{R}^d} \int_{\mathbb{R}^d} \left[q_{Z_{\tau}^{\bar x}|Z_{0}^{\bar x}}({z}_{\tau}^{x} | \bar{z}_{0}^{\bar x}) \, \psi_{(x,\,\nu^2\mathbf{I}_d)}(\bar{x})\right]\, q_{X,Z_0}(\bar{x},\bar{z}_0^{\bar{x}})\, d \bar{z}^{\bar x}_0 d \bar{x}},
\end{equation}
satisfying that $\int_{\mathbb{R}^d}\int_{\mathbb{R}^d} \widehat{w}({z}_{\tau}^{x},  z_{0}^{\tilde x}) q_{X,Z_0}(\tilde{x}, z_{0}^{\tilde x}) d{z}_{0}^{\tilde x} d\tilde{x} = 1$.  

Note that we have not yet specify the prior distribution $p_X(\tilde x)$ in Eq.~\eqref{eq:bayes}. In this work, we assume that the samples $\{x_m\}_{m=1}^M$ of $\mathcal{D}_{\rm obs}$ are samples from the prior distribution $p_X(\tilde x)$, then $\{(x_m, \Delta x_m)\}_{m=1}^M$ are samples from the joint distribution defined in Eq.~\eqref{eq:jointpdf}. Therefore, we can use the samples in $\mathcal{D}_{\rm obs}$
to define MC estimator the score function $\widehat{S}(z_{\tau}^x, \tau)$ in Eq.~\eqref{eq:appscore} as follows:
\begin{equation}\label{eq:appscore2}
    \widehat{S}^{\rm MC}(z_{\tau}^x, \tau) := \frac{1}{M}\sum_{m=1}^M - \frac{z_{\tau}^{x}- \alpha_\tau z_{0}^{x_m}}{\beta^2_\tau} \widehat{w}^{\rm MC}({z}_{\tau}^{x},  z_{0}^{x_m}),
\end{equation}
where the weights $\widehat{w}^{\rm MC}({z}_{\tau}^{x_m},  z_{0}^{x_m})$ is defined by
\begin{equation}\label{eq:wbar}
    \widehat{w}^{\rm MC}(z_\tau^{x}, z_0^{x_m}) := \dfrac{\displaystyle \exp \left\{ \frac{- (z_\tau^{x} -  \alpha_\tau \Delta x_{m})^2}{2 \beta^2_\tau } \right\} \exp \left\{ - \frac{\| x -  x_{m} \|_2^2}{2\nu^2} \right\}   }
  {\displaystyle \sum_{m'=1}^{M} \exp \left\{ \dfrac{- (z_\tau^{x} -  \alpha_\tau \Delta x_{m'})^2}{2 \beta^2_\tau } \right\} \exp \left\{ - \frac{\| x -  x_{m'} \|_2^2}{2\nu^2}  \right\}   },
\end{equation}
for $z_0^{x_m} = \Delta x_m$ and $m = 1, \ldots, M$. 
We observe that 
$\exp \{ - {\| x -  x_{m} \|_2^2}/(2\nu^2)\}$ in Eq.~\eqref{eq:wbar} determines how much contribution each sample in $\mathcal{D}_{\rm obs}$ makes to the calculation of the approximate score function $\widehat{S}^{\rm MC}$. The expression in Eq.~\eqref{eq:appscore2} and Eq.~\eqref{eq:wbar} are mathematically rigorous in the sense that $\widehat{S}^{\rm MC}$ converges to the exact score as $\nu \rightarrow 0$ and $M \rightarrow \infty$. However, Eq.~\eqref{eq:appscore2} and Eq.~\eqref{eq:wbar} may be computationally expensive, especially when $M$ is very large.
Therefore, in the numerical experiments in Section \ref{sec:results}, we select a subset\footnote{The size of the subset is set to 1\% of the total number of samples $M$.} of $\mathcal{D}_{\rm obs}$ that contains the closest neighboring samples around $x$ and only compute the weight $\exp \{ - {\| x -  x_{m} \|_2^2}/(2\nu^2)\}$ for the subset with $\nu =1$. In this way, a much smaller number of samples are involved in the computation of the MC estimator $\widehat{S}^{\rm MC}$. Moreover, because $\exp \{ - {\| x -  x_{m} \|_2^2}/(2\nu^2)\}$ is independent of the pseudo-time $\tau$, we do not need to update the subset during solving the reverse SDE.

There are several advantages of proposed training-free score estimators compared to learning the score function using neural networks. First, our method does not require solving the forward SDE in Eq.~\eqref{eq:forward} and storing a large number of trajectories, because it does not need to use the trajectory data to train a neural network to learn the score function. Instead, our method can directly solve the reverse SDE in Eq.~\eqref{eq:backward} and use the MC estimators to approximate the score function at any state $z_\tau^x$. Second, it is easy to parallelize the solution process of the reverse SDE on a large number of GPUs. In our previous work \cite{Diff_UQ_HPC}, we implemented a parallel-in-time method to solve the reverse process of the unconditional version of our diffusion model, but the parallelization approach can be easily extended to the conditional setting.

\subsection{Supervised learning of the generative model using labeled data}\label{sec:learning}
In this section, we describe how to leverage the score function estimated in Section \ref{sec:mc} to generate labeled data for the target stochastic flow map $F_{\Delta t}( x,\omega)$ in Eq.~\eqref{eq:truefm}, such that we can train the desired generator $G_{\theta}$ in Eq.~\eqref{eq:gf} using supervised learning. We observe that the flow map $G_{\theta}(x,z)$ is a deterministic map from the terminal state $Z_1^x$ to the initial state $Z_0^x$. However, the stochastic nature of the reverse SDE in Eq.~\eqref{eq:backward} leads to a random map from $Z_1^x$ to $Z_0^x$. This means the reverse SDE cannot be used to generated labeled data for supervised training of $G_{\theta}(x,z)$. Fortunately, we can convert the reverse SDE to a reverse ODE using the property that 
\begin{equation}\label{eq:nabla}
   \nabla q_{Z_\tau^x}(z_\tau^x) = q_{Z_\tau^x}(z_\tau^x) \nabla \log(q_{Z_\tau^x}(z_\tau^x)) = q_{Z_\tau^x}(z_\tau^x) S(z_\tau^x, \tau),
\end{equation}
which holds true for all distribution of the exponential family. 
Specifically, the PDF $q_{Z_\tau^x}(z_\tau^x)$ of the reverse SDE's state is the solution of a reverse Fokker-Planck equation:
\begin{equation}\label{eq:FP}
    \frac{\partial q_{Z_\tau^x}(z_\tau^x)}{\partial \tau} =  \nabla \cdot \left[ (b(\tau) z_{\tau}^x - \sigma^2(\tau) S(z_\tau^x,\tau))q_{Z_\tau^x}(z_\tau^x) + \frac{\sigma^2(\tau)}{2} \nabla q_{Z_\tau^x}(z_\tau^x)\right], 
\end{equation}
where $b(\tau)$, $\sigma(\tau)$ and $S(z_\tau^x,\tau)$ are the same as in Eq.~\eqref{eq:backward}. Substituting Eq.~\eqref{eq:nabla} into Eq.~\eqref{eq:FP}, we can convert the reverse Fokker-Planck equation to a reverse convection equation, i.e.,
\begin{equation}\label{eq:convection}
\begin{aligned}
        \frac{\partial q_{Z_\tau^x}(z_\tau^x)}{\partial \tau} & = \nabla \cdot 
        \left[ g(z_\tau^x,\tau) q_{Z_\tau^x}(z_\tau^x)\right],\\
        g(z_\tau^x,\tau)& =   (b(\tau) z_{\tau}^x - \sigma^2(\tau) S(z_\tau^x,\tau)) + \frac{\sigma^2(\tau)}{2} \nabla \log(q_{Z_\tau^x}(z_\tau^x))\\
        & = b(\tau) z_{\tau}^x - \frac{\sigma^2(\tau)}{2} S(z_\tau^x,\tau), 
\end{aligned}
\end{equation}
which corresponds to the following reverse ODE 
\begin{equation}\label{eq:revODE}
d{Z}_\tau^x = \left[ b(\tau){Z}_\tau^x - \frac{1}{2}\sigma^2(\tau) S(Z_\tau^x, \tau)\right] d\tau,  \qquad \text{with}\; Z_1^x =  Z \sim \mathcal{N}(0, \mathbf{I}_d),
\end{equation}
whose state has the same distribution as the reverse SDE. In addition, this ODE has a unique solution and thus provides a smoother function relationship between the initial state $Z_0^x $ and the terminal state $Z_1^x $. 
Therefore, we adopt the ODE in Eq.~\eqref{eq:revODE} to generate labeled data for supervised learning of the flow map $F_{\Delta t}(x, \omega)$ in Eq.~\eqref{eq:truefm}.
In this work we use the simple Euler scheme to solve the reverse ODE, i.e., 
\begin{equation}\label{eq:euler}
    z_{\tau_{k-1}}^x = z_{\tau_{k}}^x - \left[ b(\tau_k){z}_{\tau_k}^x - \frac{1}{2}\sigma^2(\tau_k) S(z_{\tau_k}^x, \tau_k)\right] \Delta \tau, \qquad k = K, \ldots, 1,
\end{equation}
where $\Delta \tau = 1/K$ and $\tau_k = k\Delta \tau$ for $k = 0, \ldots, K$.
We denote the labeled data by 
\begin{equation}\label{eq:label}
    \mathcal{D}_{\rm label} := \left\{ (x_j, z_j, y_j)\, \big |\, x_j \in \mathcal{D}_{\rm obs} \text{ and } z_j \sim \mathcal{N}(0,\mathbf{I}_d), \; j = 1, \ldots, J \right\},
\end{equation}
where $J\leq M$ is the size of the labeled data set, $x_j$ is a sample from $\{x_m\}_{m=1}^M \subset \mathcal{D}_{\rm obs}$, $z_j$ is a sample from the 
standard normal distribution $\mathcal{N}(0,\mathbf{I}_d)$, and $y_j$ is the generated labeled data by solving the reverse ODE using the training-free score estimator in Section \ref{sec:mc}. For $j=1, \ldots, J$, we solve the reverse ODE from $\tau = 1$ to $\tau=0$ by setting $x = x_j$ and $z_1^{x_j} = z_j$ in Eq.~\eqref{eq:revODE}.
After solving the ODE, we collect the state $z_0^{x_j}$ and let $y_j = z_0^{x_j}$. 
Once the entire labeled dataset $\mathcal{D}_{\rm label}$ is generated, we train a fully-connected neural network $G_\theta(x,z)$ to approximate the flow map $F_{\Delta t}( x, \omega)$ in Eq.~(\ref{eq:truefm}) using the standard mean squared error (MSE) loss. The workflow of the proposed method is summarized in Algorithm \ref{alg:cap}. 
\begin{algorithm}[h!]
\caption{The training-free conditional diffusion model}\label{alg:cap}
\setstretch{1.2}
\begin{algorithmic}
\State {\bf Input:} Observation data set $\mathcal{D}_{\rm obs} = \{(x_m, \Delta x_m)\}_{m=1}^M$
\State {\bf Procedure:}
\For{$j = 1, 2, \ldots, J$}\Comment{\texttt{Generating labeled data}}
\State Random select a sample $x_j$ from $\{x_m\}_{m=1}^M \subset \mathcal{D}_{\rm obs}$;
\State Generate a sample $z_j$ from $\mathcal{N}(0, \mathbf{I}_d)$ and set $z_{\tau_K}^{x_j} = z_j$ in Eq.~\eqref{eq:euler};
\For{$k = K,\ldots, 1$}
\State Compute the weight $\{\widehat{w}^{\rm MC}(z_{\tau_k}^{x_j}, z_0^{x_m})\}_{m=1}^M$ using Eq.~\eqref{eq:wbar};
\State Compute the score function $\widehat{S}^{\rm MC}(z_{\tau_k}^{x_j}, \tau_k)$ using Eq.~\eqref{eq:appscore2};
\State Solve $z_{\tau_{k-1}}^{x_j}$ using the Euler scheme in Eq.~\eqref{eq:euler}
\EndFor
\State Set $y_j = z_{\tau_0}^{x_j}$ and assemble $(x_j, z_j, y_j)$
\EndFor
\State Train $G_\theta(x,z)$ using the labeled data $\mathcal{D}_{\rm label} = \{(x_j,z_j,y_j)\}_{j=1}^J$;
\State {\bf Output:} The trained generative model $G_\theta(x,z)$.
\end{algorithmic}
\end{algorithm}

\section{Numerical experiments}\label{sec:results}
In this section, we present several numerical examples to demonstrate the accuracy and efficiency of the proposed training-free diffusion model for learning flow map of stochastic dynamical systems. The examples include:
\vspace{0.2cm}
\begin{itemize}[leftmargin=15pt]\itemsep0.1cm
    \item Linear SDEs: Ornstein-Ulenbeck (OU) process and geometric Brownian motion;
    \item Nonlinear SDEs: SDEs with exponential and trigonometric drift or diffusion, as well as SDE featuring a double-well potential;
    \item SDEs with non-Gaussian noise of exponential and lognormal distributions;
    \item Multi-dimensional SDEs: 2-dimensional and 5-dimensional OU processes.
\end{itemize}
\vspace{0.1cm}

\begin{remark}[Reproducibility]
   Our method is implemented using PyTorch with GPU acceleration. The source code has been made publicly accessible at \url{https://github.com/YanfangLiu11/Conditional-Diffusion-Model-for-SDE-Learning}. All numerical results presented in this section are fully reproducible using the provided GitHub repository.  
\end{remark}

{\bf Training dataset $\mathcal{D}_{\rm obs}$ generation}: For each example, we generate $H$ trajectories by solving the exact SDEs using Euler-Maruyama method, with initial conditions uniformly distributed within specified regions. These trajectories are then reorganized as data pairs $\{(x_m, \Delta x_m)\}$ as described in Section \ref{sec:problem} to form observation dataset $\mathcal{D}_{\rm obs}$. The time step in the Euler-Maruyama method is set to $\Delta t=0.01$ up to $T=1.0$ for all examples (except for geometric Brownian motion with $T=0.5$). Then, each trajectory will contribute $L=100$ (for geometric Brownian motion $L=50$) samples to $\mathcal{D}_{\rm obs}$. Accordingly, the size of observation data for each example is equal to $M= H\times L$ as shown in Eq.~\eqref{eq:obs}.
For the training-free diffusion model, the reverse-time ODE in Eq.~\eqref{eq:revODE} is solved using the explicit Euler scheme with 10,000 time steps, i.e., $K = 10,000$ in Eq.~\eqref{eq:euler}. Each ODE solution generates a labeled data $(x_{j},z_j, {y}_{j})$ for the labeled dataset $\mathcal{D}_{\rm label}$. 
This labeled dataset $\mathcal{D}_{\rm obs}$ is then divided into training and validation sets, with $80\%$ of the data for training and the remaining $20\%$ for validation. 

{\bf Supervised training of $G_\theta(x,z)$}: We define the generative model $G_\theta(x,z)$ as a fully-connected feed-forward neural network with one hidden layer to learn the flow map $F_{\Delta t}(x, \omega)$ in Eq.~\eqref{eq:truefm}. Because we have labeled data $\mathcal{D}_{\rm label}$, we use the standard Mean Squared Error (MSE) loss to train the generative model $G_\theta(x,z)$. For each example, the neural network is trained for 2,000 epochs using the Adam optimizer with a learning rate of 0.01. During the training period, the best neural network parameters are saved based on validation performance to ensure optimal training loss. Since the neural network structure is simple, we use the simple grid search method to tune the number of neurons in the hidden layer. 

{\bf Metrics for testing the generative model $G_\theta(x,z)$}:
The trained conditional generative model $G_\theta(x,z)$ is applied to predict each target SDE system for a time horizon typically up to $T=5 \sim 500$, which is much longer time horizon that the time horizon covered by the observation data $\mathcal{D}_{\rm obs}$ where the time horizon is $T=1$ or $0.5$. For all the numerical examples, we simulate 500,000 trajectories using both the approximate flow map and the ground truth flow map. The prediction is then compared with the ground truth solution using the following metrics:
\vspace{0.2cm}
\begin{itemize}[leftmargin=15pt]\itemsep0.0cm
    \item The mean and standard deviation of the simulated trajectories for visual comparison of the accuracy;
    \item Comparison of the approximate drift and diffusion coefficients, obtained from simulated trajectories, against the true drift and diffusion coefficients;
    \item Comparison of the approximate flow map $G_\theta$ to the ground truth flow map $F_\Delta$;
\end{itemize}
\vspace{0.2cm}

\subsection{Linear SDEs} We first consider the learning of an Ornstein-Uhlenbeck (OU) process and a geometric Brownian monition.

\subsubsection{One-dimensional OU process} The one-dimensional OU process under consideration is defined by
\begin{equation}\label{eq: OU_SDE}
    dX_t = \theta (\mu-X_t) dt +\sigma dW_{t},
\end{equation}
where $\theta = 1.0$, $\mu=1.2$ and $\sigma=0.3$. The observation dataset $\mathcal{D}_{\rm obs}$ consists of $M=1,500K$ data pairs from $H=15,000$ trajectories, obtained by solving the SDE in \eqref{eq: OU_SDE} with initial values uniformly sampled from $\mathcal{U}(0,2.5)$ up to the time horizon $T=1.0$. We randomly choose $50,000$ initial states from sample $\mathcal{D}_{\rm obs}$ to generate the corresponding labeled training data. After training the generative model $G_\theta$, we simulate 500,000 prediction trajectories for the time horizon up to $T=5.0$.

\begin{figure}[h!]
    \centering
    \includegraphics[width=0.6\textwidth]{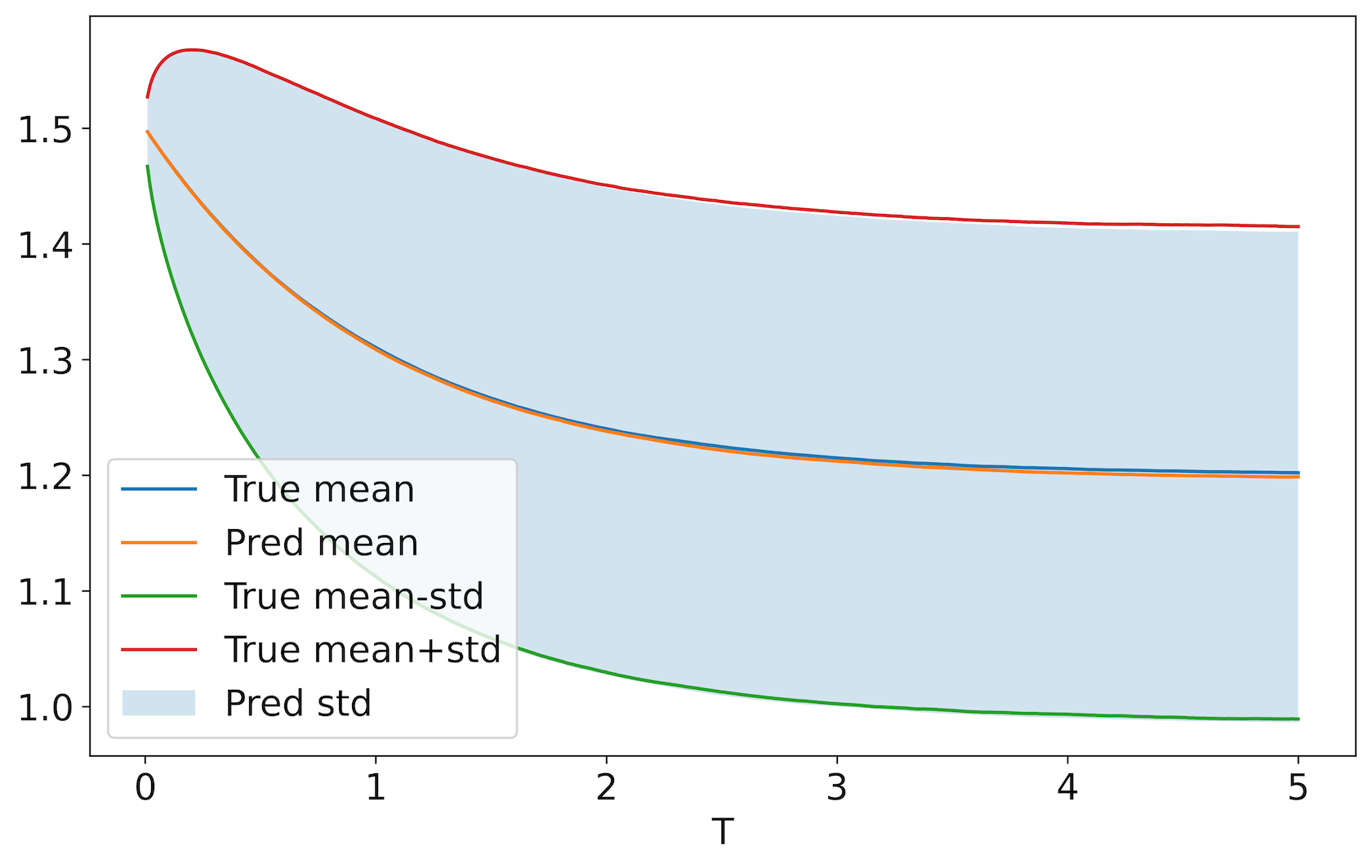}\vspace{-0.15in}
    \caption{One-dimensional OU process: comparison of the mean and the standard deviation of solutions with the initial state being $X_0=1.5$, obtained by the generative model and the ground truth.}
    \label{fig:OU_mean_std}
\end{figure}

\begin{figure}[h!]
    \centering
    \includegraphics[width=0.8\textwidth]{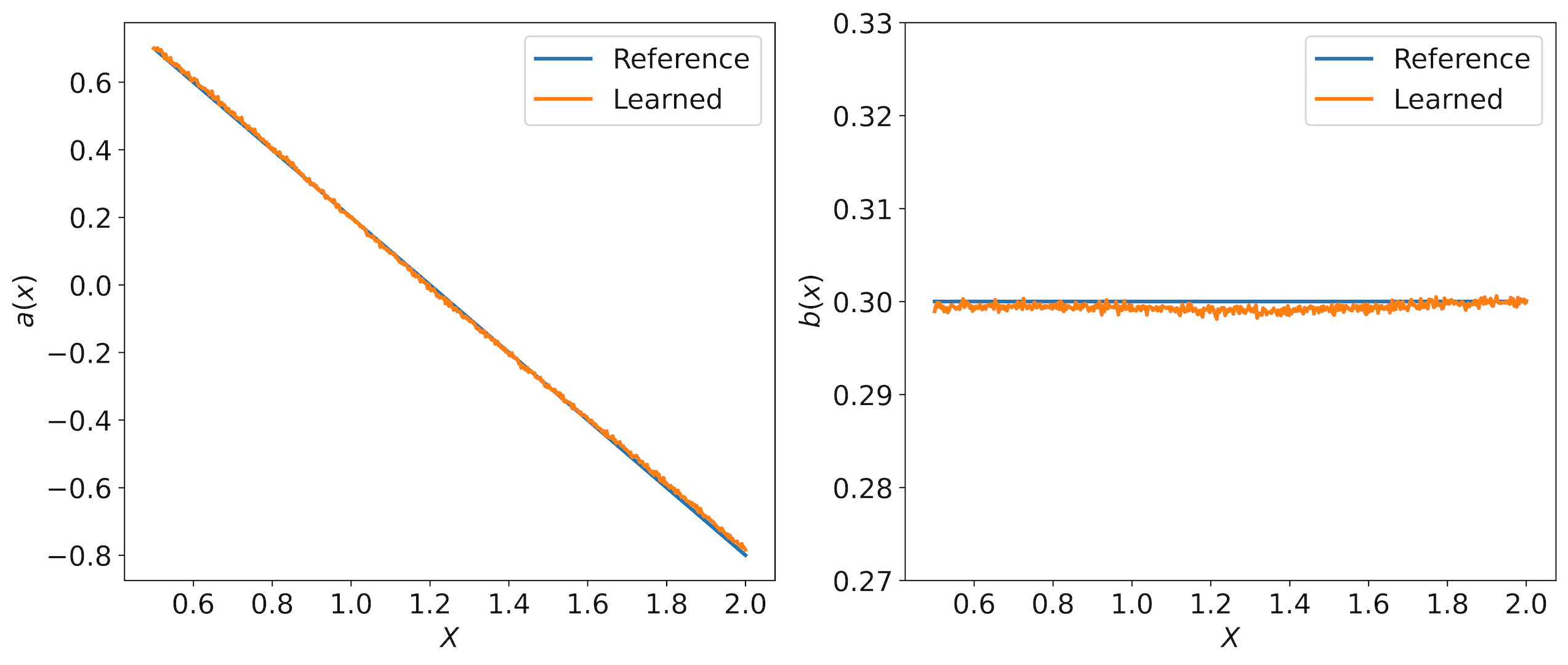} \vspace{-0.1in}
    \caption{One-dimensional OU process: comparison of effective drift and diffusion functions obtained by the simulated trajectories using the generative model and the exact SDE. Left: drift $a(x)= \mu-x$; Right: diffusion $b(x)=\sigma$.}
    \label{fig:OU_drift_diff}
\end{figure}

\begin{figure}[!ht]
    \centering
    \includegraphics[width=0.6\textwidth]{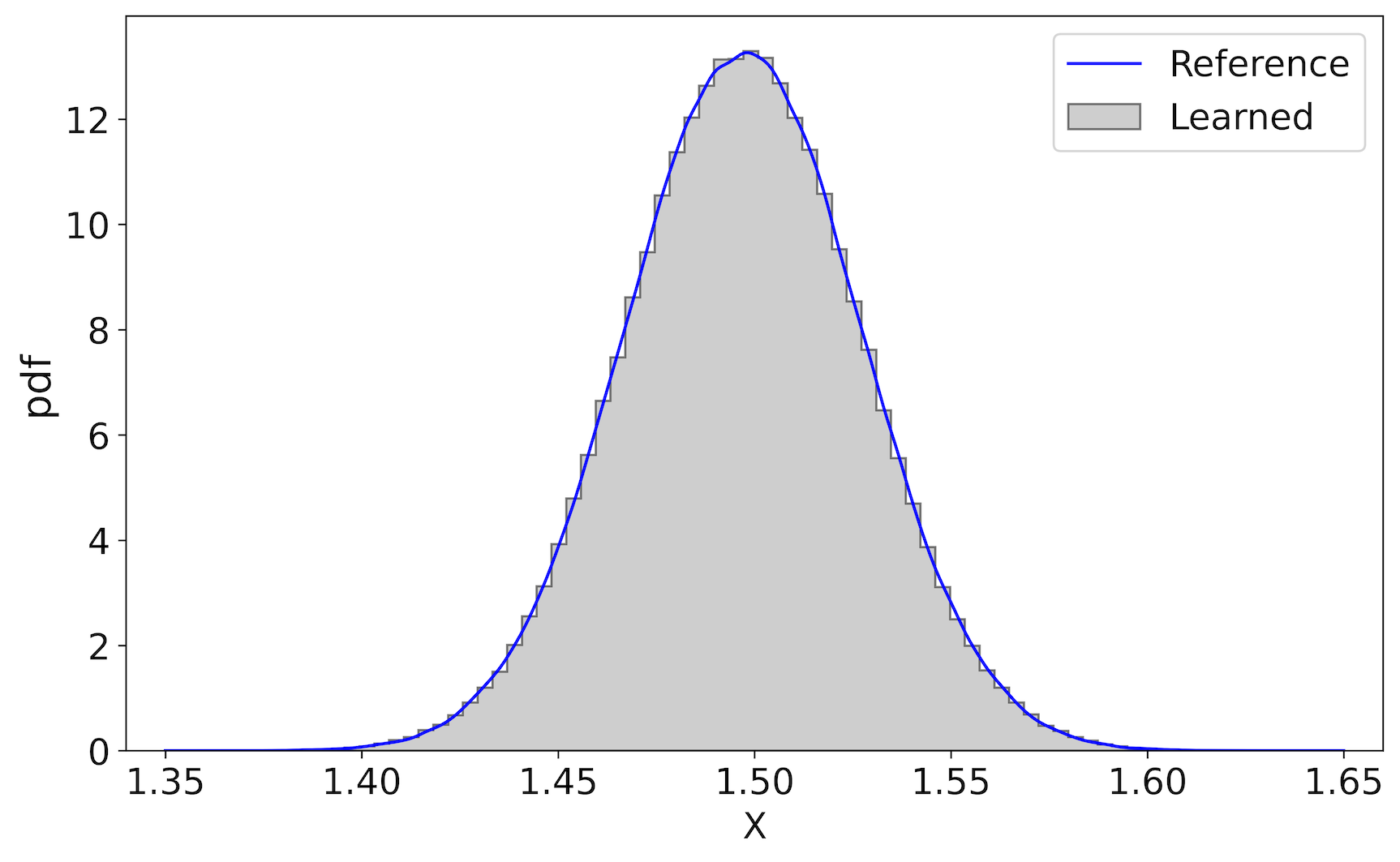}\vspace{-0.15in}
    \caption{One-dimensional OU process: comparison of conditional PDF $p_{X_{t+\Delta t}|X_t}(x_{t+\Delta t}|x_{t}=1.5)$ determined by the generative model $G_\theta$ and the exact flow map $F_{\Delta t}$. }
    \label{fig:OU_pdf}
\end{figure}

The mean and standard deviation of the predicted solutions when $X_{0}=1.5$ by the generative model, compared with those of the exact solutions are displayed in Figure \ref{fig:OU_mean_std}. Good agreements between the mean and standard deviation of the predicted solutions and that of the exact solution can be observed for time up to $T=5.0$ despite training dataset being limited to $T=1.0$. In Figure \ref{fig:OU_drift_diff}, we present the effective drift and diffusion determined by the generative model, which are closely aligned with the true drift and diffusion. The relative errors for drift and diffusion functions are on the order of $10^{-2}$ and $10^{-3}$, respectively. Figure \ref{fig:OU_pdf} illustrates the one-step conditional distribution $p_{X_{t+\Delta t}|X_t}(x_{t+\Delta t}|x_{t}=1.5)$ determined by the generative model $G_\theta$ and exact flow map $F_{\Delta t}$. The predicted conditional distribution accurately approximates the exact one.


\subsubsection{Geometric Brownian Motion} The one-dimensional geometric Brownian motion is defined by
\begin{equation}\label{eq: GBM_SDE}
    dX_t = \mu X_t dt +\sigma X_t dW_{t},
\end{equation}
where $\mu = 2$, $\sigma=1$. The observation dataset $\mathcal{D}_{\rm obs}$ consists of $M=5,000K$ data pairs from $H=100,000$ trajectories of the SDE, obtained by solving the SDE in Eq.~\eqref{eq: GBM_SDE} with initial values uniformly sampled from $\mathcal{U}(0,2)$ up to $T=0.5$. We randomly choose $120,000$ initial states from sample $\mathcal{D}_{\rm obs}$ to generate the corresponding labeled training data. After training the generative model $G_\theta$, the prediction trajectories are simulated over a time horizon up to $T=1.0$.

\begin{figure}[!ht]
    \centering
    \includegraphics[width=0.6\textwidth]{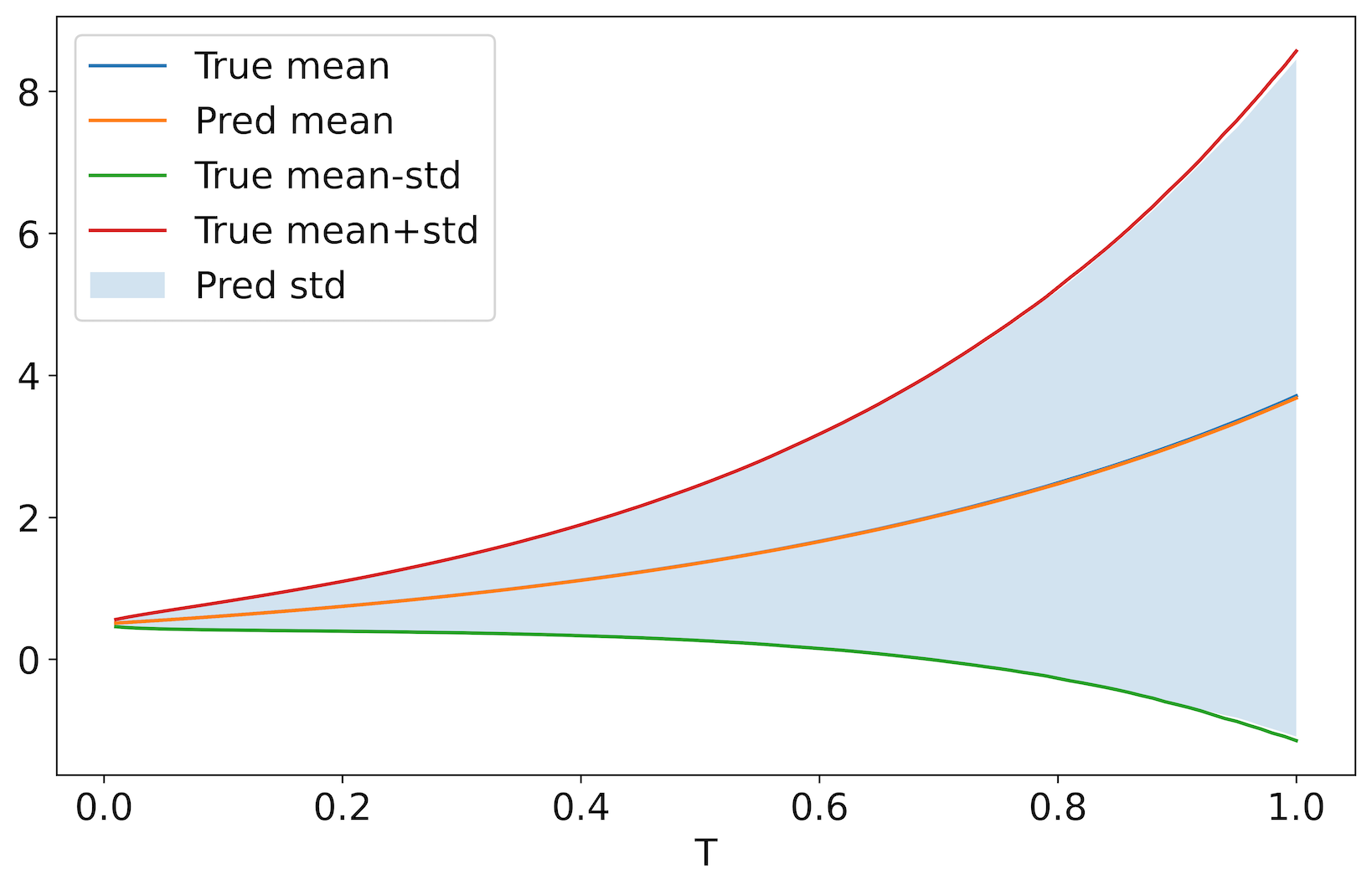}\vspace{-0.15in}
    \caption{One-dimensional GBM: comparison of the mean and the standard deviation of solutions with the initial state being $X_0=0.5$, obtained by the generative model and the ground truth.}
    \label{fig:GBM_mean_std}
\end{figure}

\begin{figure}[!ht]
    \centering
    \includegraphics[width=0.8\textwidth]{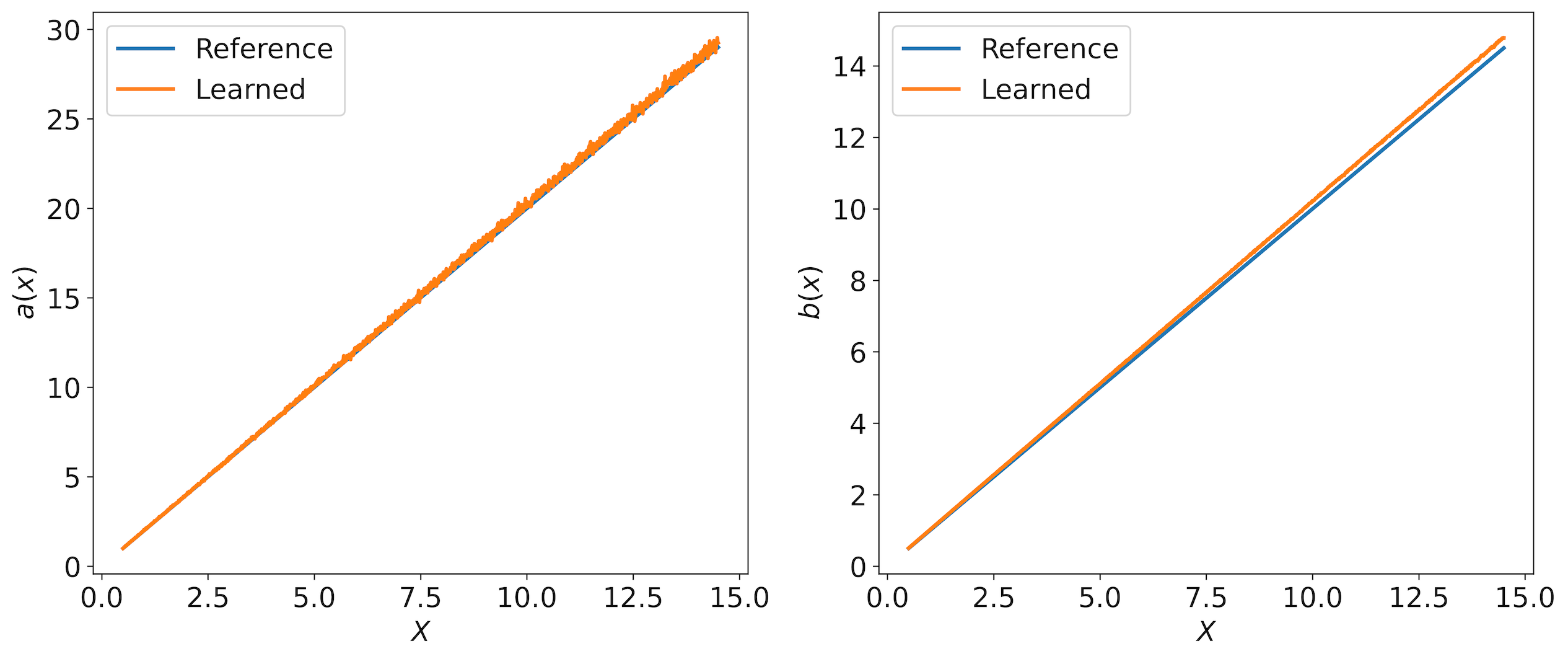} \vspace{-0.1in}
    \caption{One-dimensional GBM: comparison of effective drift and diffusion functions obtained by the simulated trajectories using the generative model and the exact SDE. Left: drift $a(x)= \mu x$; Right: diffusion $b(x)=\sigma x$.}
    \label{fig:GBM_drift_diff}
\end{figure}

\begin{figure}[!ht]
    \centering
    \includegraphics[width=0.6\textwidth]{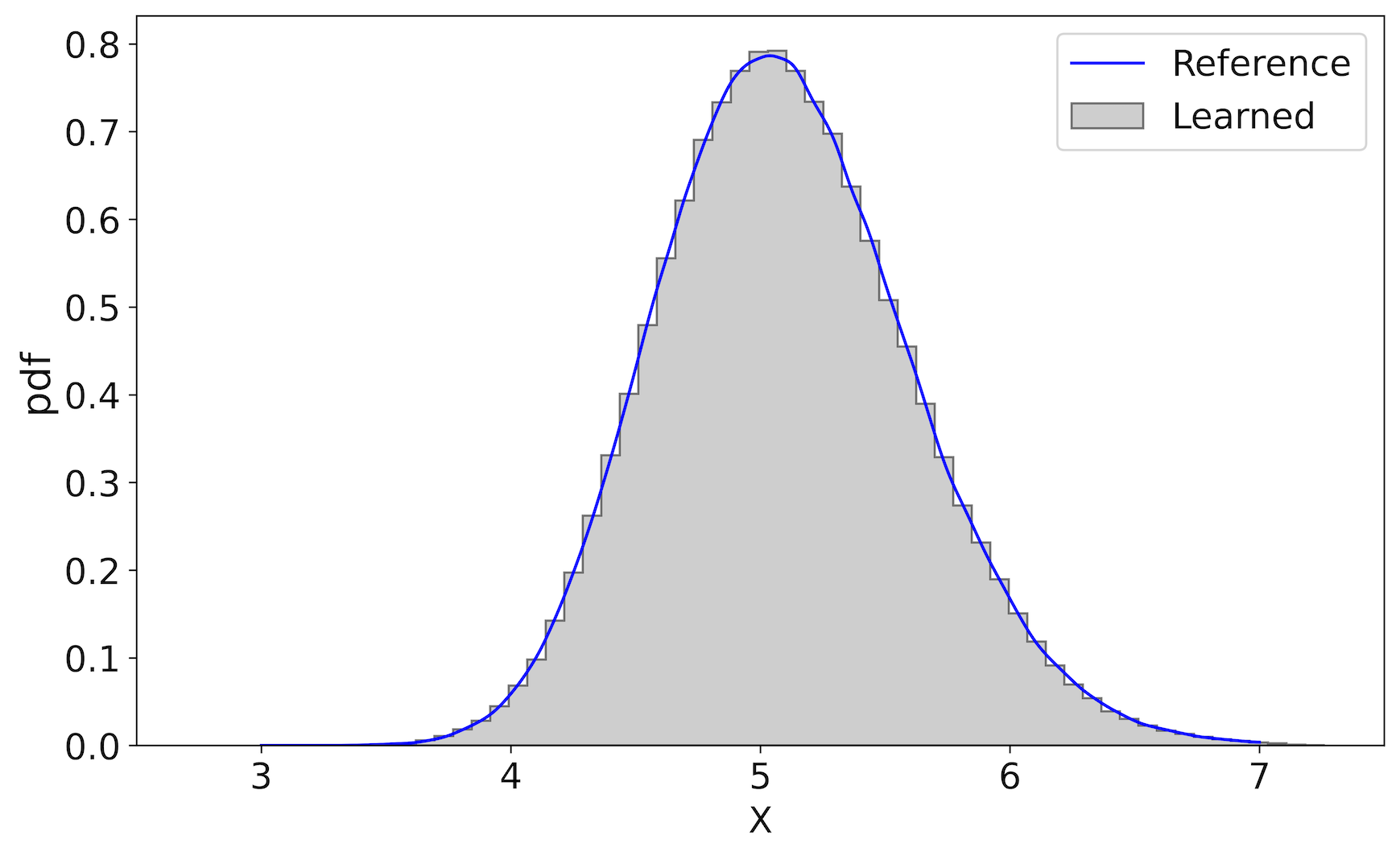}\vspace{-0.15in}
    \caption{One-dimensional GBM: comparison of conditional PDF $p_{X_{t+\Delta t}|X_t}(x_{t+\Delta t}|x_{t}=5.0)$ determined by the generative model $G_\theta$ and the exact flow map $F_{\Delta t}$.}
    \label{fig:GBM_pdf}
\end{figure}

The mean and standard deviation of the predicted solutions for $X_{0}=0.5$ by the generative model, compared with those of the exact solutions are displayed in Figure \ref{fig:GBM_mean_std}. Good agreements between the mean and standard deviation of the predicted solutions and that of the exact solution can be observed for time up to $T=1.0$ despite training dataset being limited to $T=0.5$. In Figure \ref{fig:GBM_drift_diff}, we present the effective drift and diffusion determined by the generative model, which are closely aligned with the true drift and diffusion. The relative errors for both drift and diffusion functions are on the order of $10^{-2}$. Figure \ref{fig:GBM_pdf} illustrates the one-step conditional distribution $p_{X_{t+\Delta t}|X_t}(x_{t+\Delta t}|x_{t}=5.0)$ for both generative model and exact SDE. The predicted conditional distribution accurately approximates the exact one.


\subsection{Nonlinear SDEs} This section considers two SDEs with exponential and trigonometric drift or diffusion functions, as well as SDEs featuring a double-well potential.
\subsubsection{SDE with nonlinear diffusion} The SDE with a non-linear diffusion is defined by
\begin{equation}\label{eq: Expdiff_SDE}
    dX_t = -\mu X_t dt +\sigma e^{-X^2_t} dW_{t},
\end{equation}
where $\mu=5$ and $\sigma = 0.5$. The observation dataset $\mathcal{D}_{\rm obs}$ consists of $M=15,000$K data pairs from $H=150,000$ trajectories, obtained by solving the SDE \eqref{eq: Expdiff_SDE} with initial values uniformly sampled from $\mathcal{U}(-1,1)$ up to $T=1.0$. We randomly choose $60,000$ initial states from $\mathcal{D}_{\rm obs}$ to generate the corresponding labeled training data. After training the generative model $G_\theta$, we simulate 500,000 prediction trajectories up to $T=10.0$.

\begin{figure}[h!]
    \centering
    \includegraphics[width=0.6\textwidth]{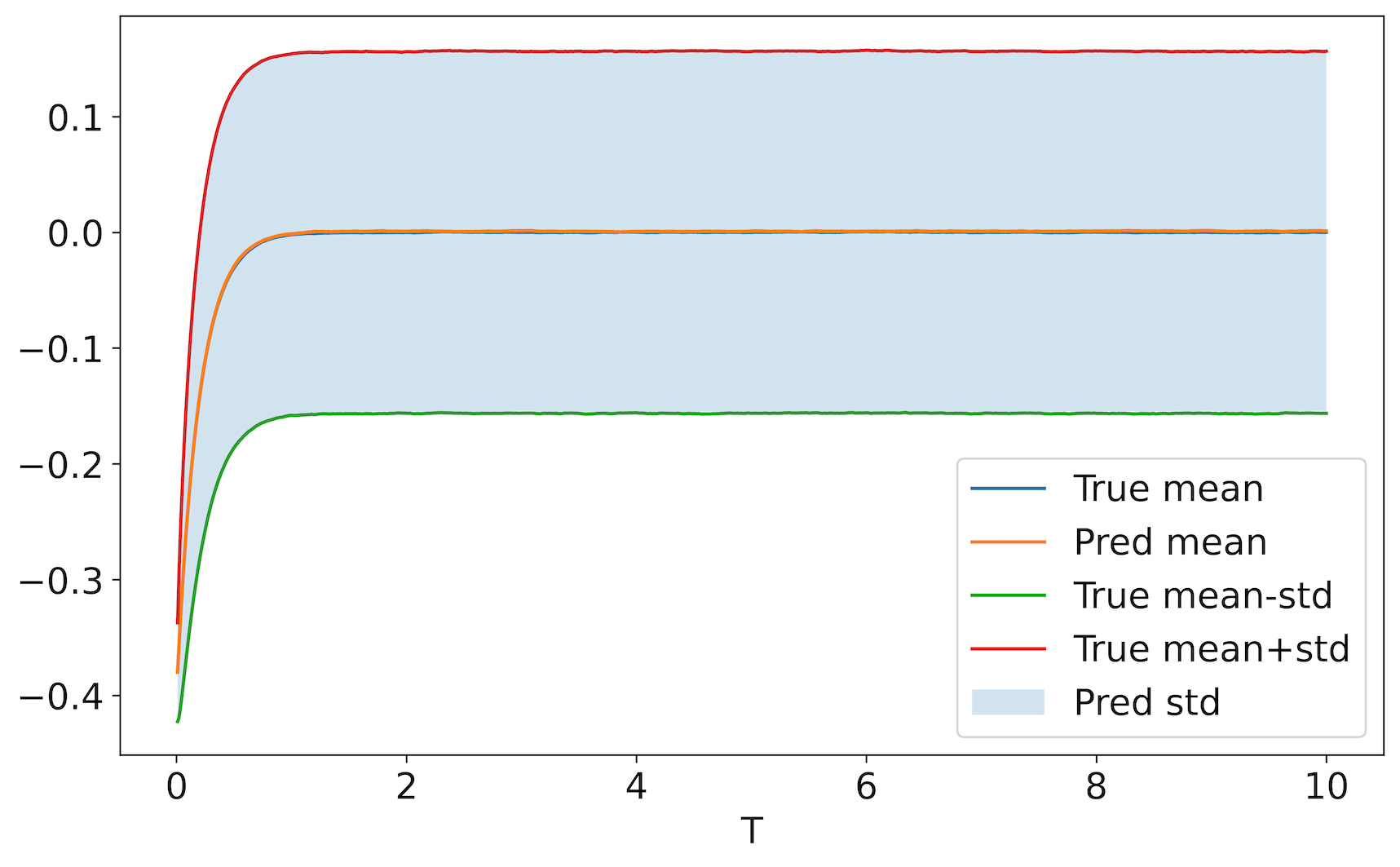}\vspace{-0.15in}
    \caption{SDE with nonlinear diffusion: comparison of the mean and the standard deviation of solutions with the initial state being $X_0=-0.4$, obtained by the generative model and the ground truth.}
    \label{fig:Expdiff_mean_std}
\end{figure}

\begin{figure}[h!]
    \centering
    \includegraphics[width=0.8\textwidth]{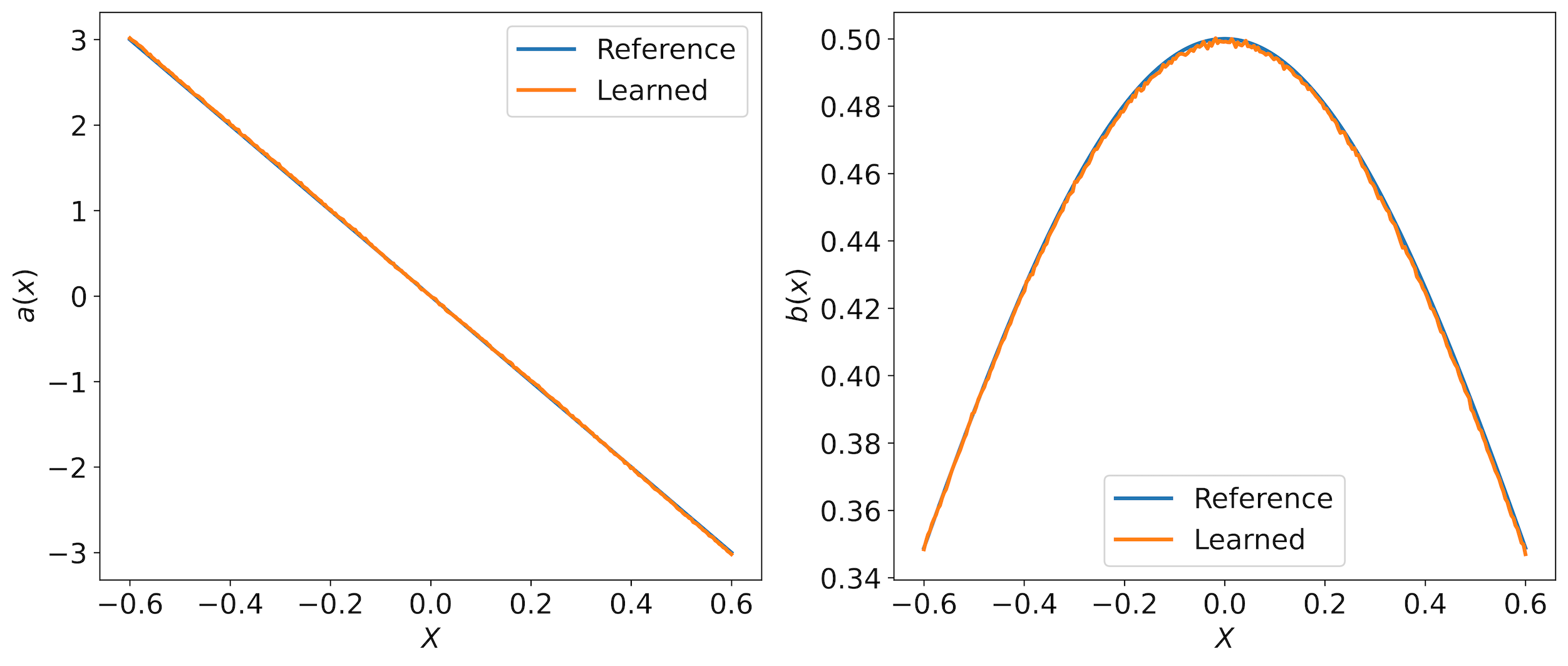} \vspace{-0.1in}
    \caption{SDE with nonlinear diffusion: comparison of effective drift and diffusion functions obtained by the simulated trajectories using the generative model and the exact SDE. Left: drift $a(x)= -\mu x$; Right: diffusion $b(x)=\sigma e^{-x^2}$.}
    \label{fig:Expdiff_drift_diff}
\end{figure}

\begin{figure}[h!]
    \centering
    \includegraphics[width=0.6\textwidth]{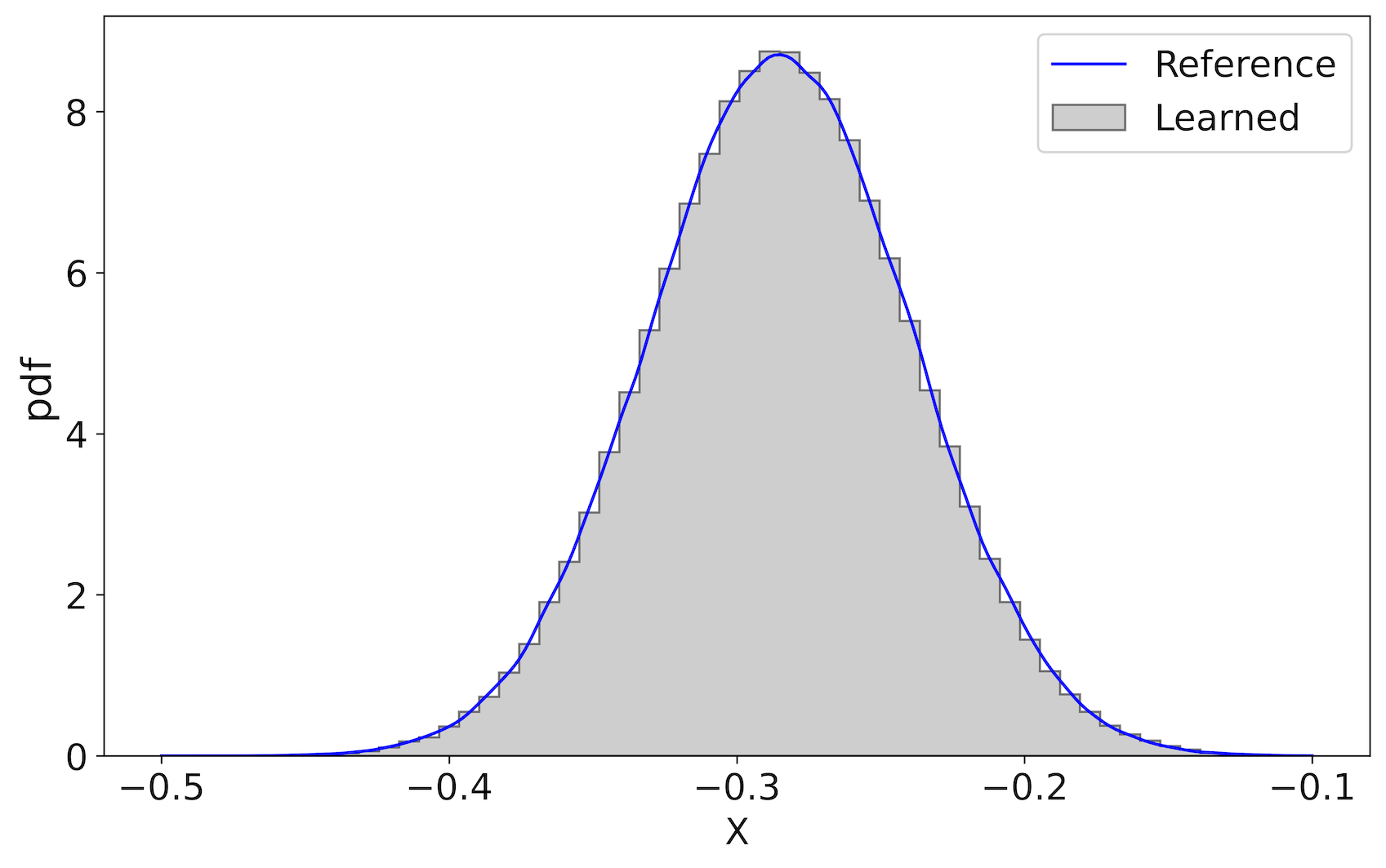}\vspace{-0.15in}
    \caption{SDE with nonlinear diffusion: comparison of conditional PDF $p_{X_{t+\Delta t}|X_t}(x_{t+\Delta t}|x_{t}=-0.3)$ determined by the generative model $G_\theta$ and the exact flow map $F_{\Delta t}$. }
    \label{fig:Expdiff_pdf}
\end{figure}

The mean and standard deviation of the predicted solutions for $X_{0}=-0.4$ by the generative model, compared with those of the exact solutions are displayed in Figure \ref{fig:Expdiff_mean_std}. Good agreements between the mean and standard deviation of the predicted solutions and that of the exact solution can be observed for time up to $T=10.0$ despite training dataset being limited to $T=1.0$. In Figure \ref{fig:Expdiff_drift_diff}, we present the effective drift and diffusion determined by the generative model, which are closely aligned with the true drift and diffusion coefficients. The relative errors for drift and diffusion functions are on the order of $10^{-2}$ and $10^{-3}$, respectively. Figure \ref{fig:Expdiff_pdf} illustrates the one-step conditional distribution $p_{X_{t+\Delta t}|X_t}(x_{t+\Delta t}|x_{t}=-0.3)$ for both generative model and exact SDE. The predicted conditional distribution accurately approximates the exact one.

\subsubsection{Trigonometric SDE} The SDE with trigonometric drift and diffusion is defined by
\begin{equation}\label{eq: Trigdiff_SDE}
    dX_t = \sin(2k\pi X_t) dt +\sigma \cos(2k\pi X_t) dW_{t},
\end{equation}
where $k=1$ and $\sigma = 0.5$. The observation dataset $\mathcal{D}_{\rm obs}$ consists of $M=20,000K$ data pairs from $H=200,000$ trajectories, obtained by solving the SDE in Eq.~\eqref{eq: Trigdiff_SDE} with initial values uniformly sampled from $\mathcal{U}(0.35,0.7)$ up to $T=1.0$. We randomly choose $60,000$ initial states from $\mathcal{D}_{\rm obs}$ to generate the corresponding labeled training data. After training the generative model $G_\theta$, we simulate 500,000 prediction trajectories up to $T=10.0$.

\begin{figure}[h!]
    \centering
    \includegraphics[width=0.6\textwidth]{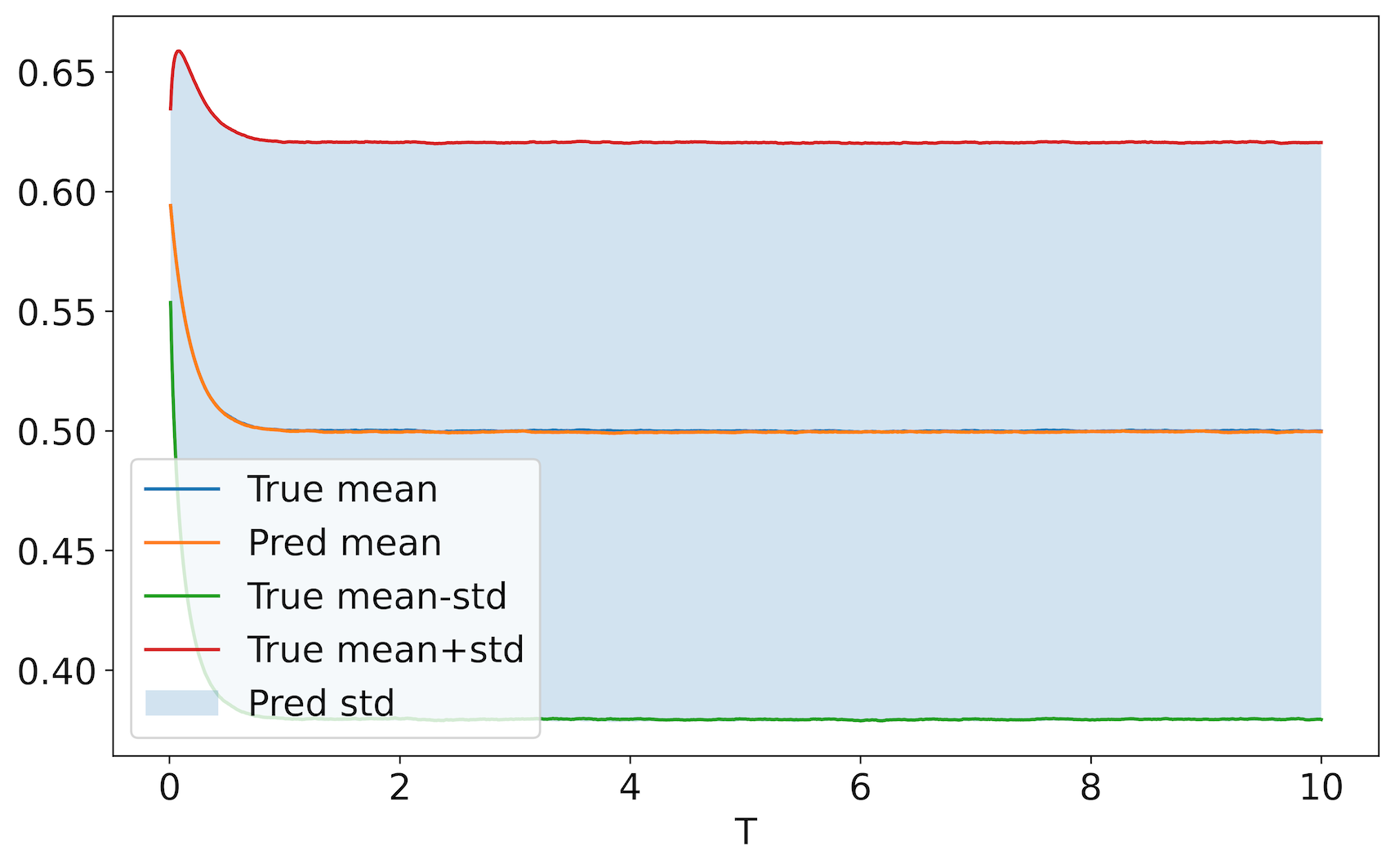}\vspace{-0.15in}
    \caption{Trigonometric SDE: comparison of the mean and the standard deviation of solutions with the initial state being $X_0=0.6$, obtained by the generative model and the ground truth.}
    \label{fig:Trigdiff_mean_std}
\end{figure}

\begin{figure}[h!]
    \centering
    \includegraphics[width=0.8\textwidth]{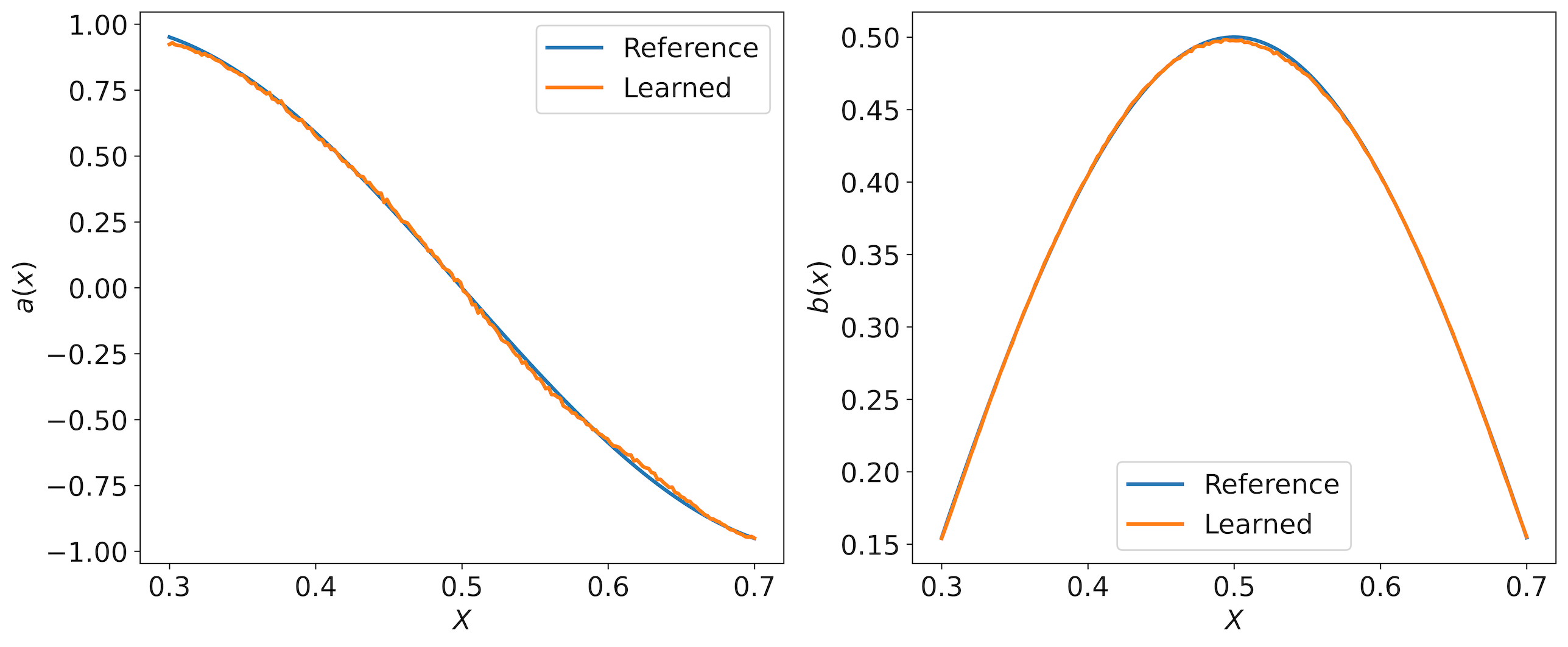} \vspace{-0.1in}
    \caption{Trigonometric SDE: comparison of effective drift and diffusion functions obtained by the simulated trajectories using the generative model and the exact SDE. Left: drift $a(x)= sin(2k\pi x)$; Right: diffusion $b(x)=\sigma\cos(2k\pi x)$.}
    \label{fig:Trigdiff_drift_diff}
\end{figure}

\begin{figure}[!ht]
    \centering
    \includegraphics[width=0.6\textwidth]{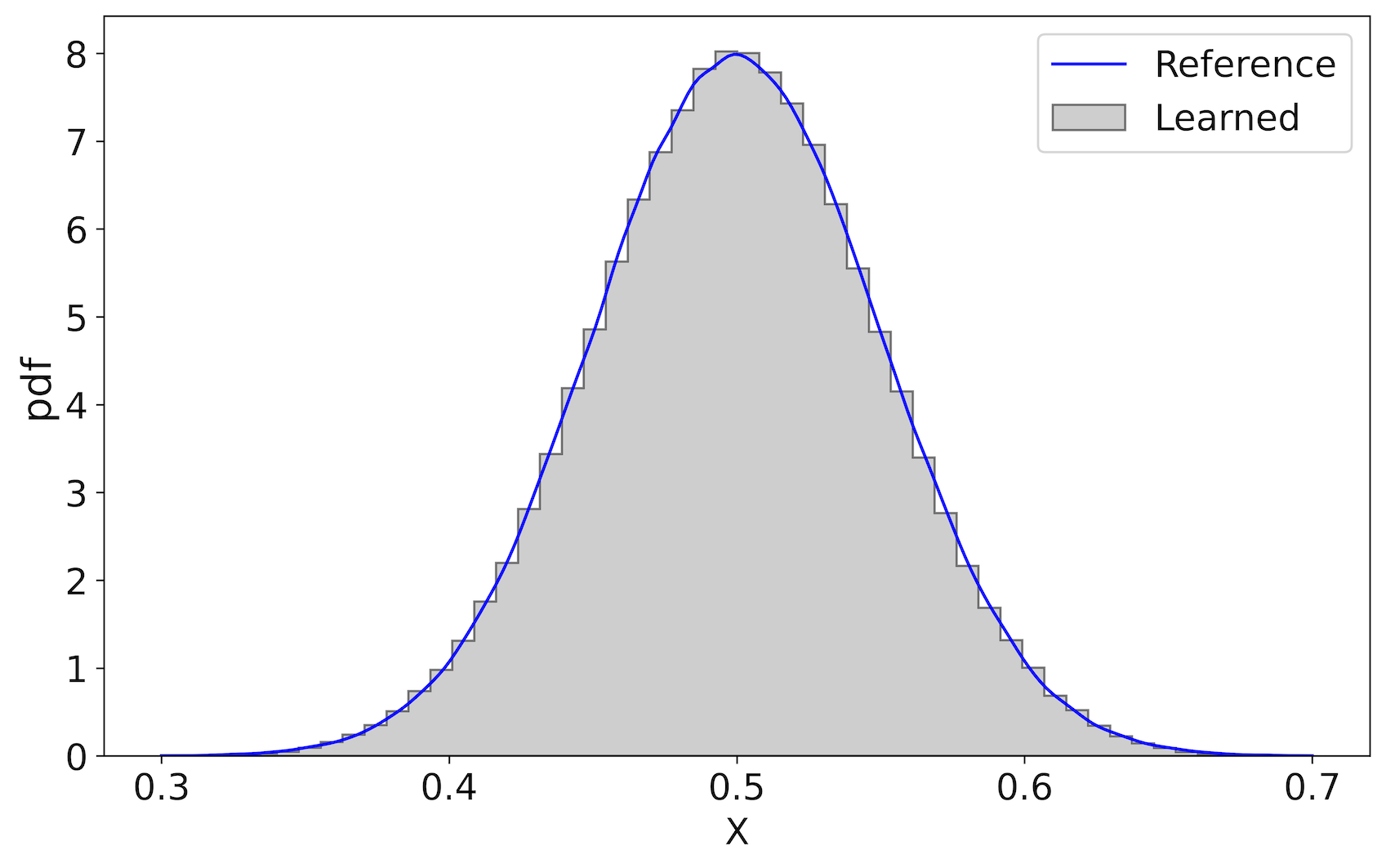}\vspace{-0.15in}
    \caption{Trigonometric SDE: comparison of conditional PDF $p_{X_{t+\Delta t}|X_t}(x_{t+\Delta t}|x_{t}=0.5)$ determined by the generative model $G_\theta$ and the exact flow map $F_{\Delta t}$.}
    \label{fig:Trigdiff_pdf}
\end{figure}

The mean and standard deviation of the predicted solutions for $X_{0}=0.6$ by the generative model, compared with those of the exact solutions are displayed in Figure \ref{fig:Trigdiff_mean_std}. Good agreements between the mean and standard deviation of the predicted solutions and that of the exact solution can be observed for time up to $T=10.0$ despite training dataset being limited to $T=1.0$. In Figure \ref{fig:Trigdiff_drift_diff}, we present the effective drift and diffusion determined by the generative model, which are closely aligned with the true drift and diffusion. The relative errors for drift and diffusion functions are on the order of $10^{-2}$ and $10^{-3}$, respectively. Figure \ref{fig:Trigdiff_pdf} illustrates the one-step conditional distribution $p_{X_{t+\Delta t}|X_t}(x_{t+\Delta t}|x_{t}=0.5)$ for both the generative model and the exact SDE. The predicted conditional distribution accurately approximates the exact one.

\subsubsection{SDE with double well potential} The SDE with a double well potential is defined by
\begin{equation}\label{eq: Dwell_SDE}
    dX_t = (X_t- X^3_t)dt +\sigma  dW_{t},
\end{equation}
where $\sigma = 0.5$. There are two stable states at $x=\pm 1$, and the system can randomly transition between these two stable states over time. The observation dataset $\mathcal{D}_{\rm obs}$ consists of $M=10,000K$ data pairs from $H=100,000$ trajectories, obtained by solving the SDE in Eq.~\eqref{eq: Dwell_SDE} with initial values uniformly sampled from $\mathcal{U}(-2.5,2.5)$ up to $T=1.0$. We randomly choose $60,000$ initial states from sample $\mathcal{D}_{\rm obs}$ to generate the corresponding labeled training data. After training the generative model $G_\theta$, we simulate 500,000 prediction trajectories up to $T=500$.

\begin{figure}[h!]
    \centering
    \includegraphics[width=0.7\textwidth]{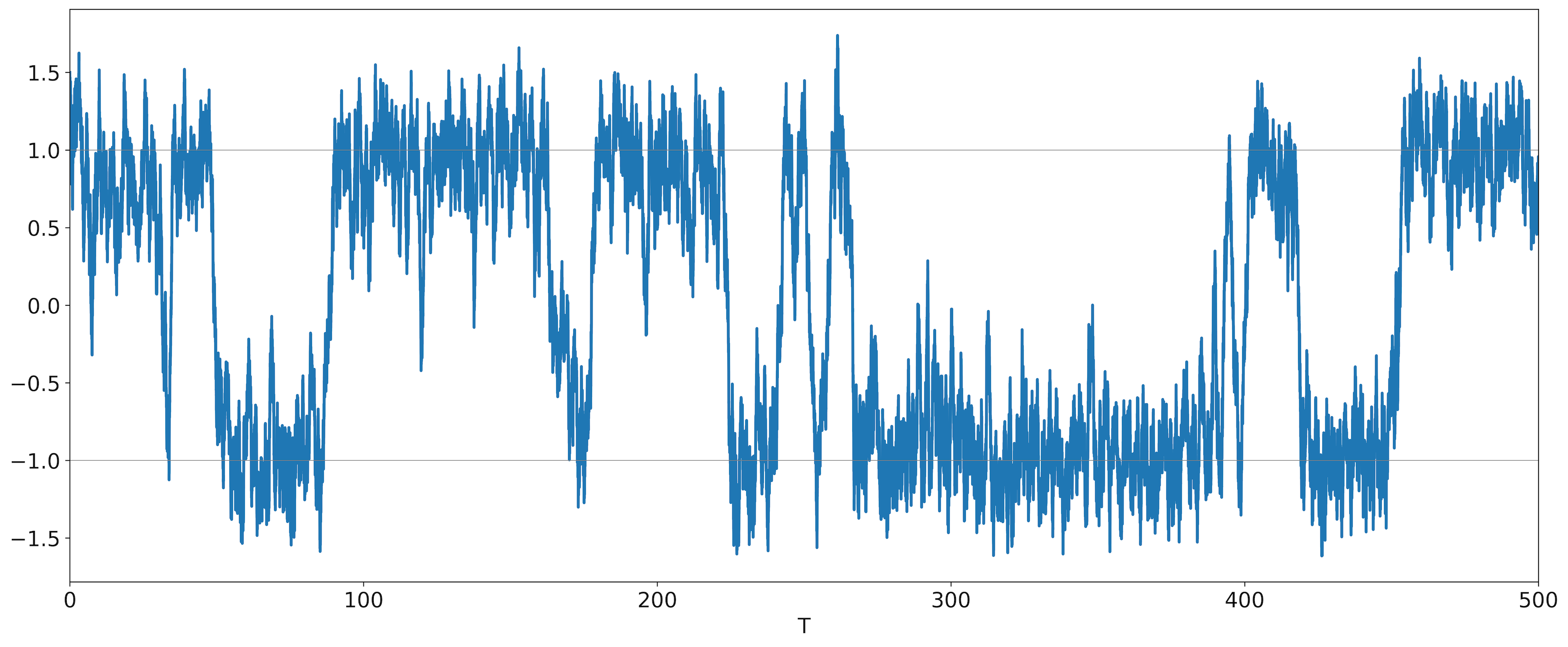}\vspace{-0.15in}
    \caption{SDE with a double well potential: solution trajectory for time up to $T=500$ when $X_{0}=1.5$.}
    \label{fig:Dwell_path}
\end{figure}

\begin{figure}[h!]
    \centering
    \includegraphics[width=0.8\textwidth]{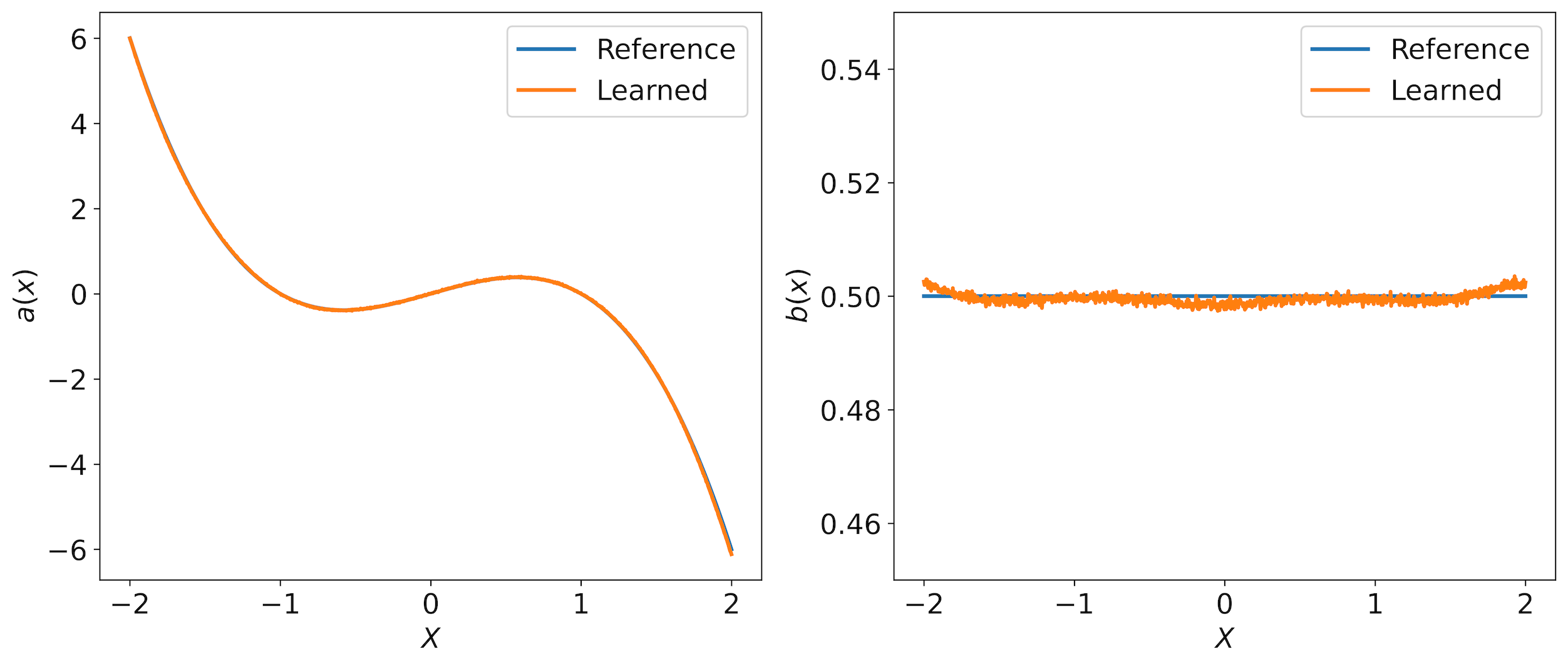} \vspace{-0.1in}
    \caption{SDE with a double well potential: comparison of effective drift and diffusion functions obtained by the simulated trajectories using the generative model and the exact SDE. Left: drift $a(x)=x-x^3$; Right: diffusion $b(x)=\sigma$.}
    \label{fig:Dwell_drift_diff}
\end{figure}

\begin{figure}[h!]
    \centering
    \includegraphics[width=\textwidth]{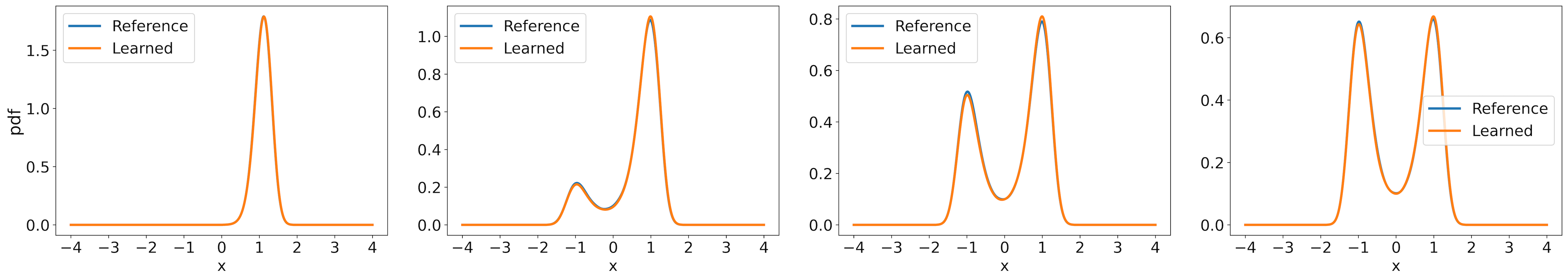} \vspace{-0.35in}
    \caption{SDE with a double well potential: temporal evolution of the solution probability distribution when $X_{0}=1.5$ at time $T=0.5$, $10$, $30$ and $100$ (from left to right).}
    \label{fig:Dwell_temporal_pdf}
\end{figure}

A predicted solution at $X_{0}=1.5$ for time up to $T=500$ by the generative model is displayed in Figure \ref{fig:Dwell_path}.  Despite the training dataset being limited to $T=1.0$, where no transitions between the two stable states are not observed, the generative model is able to predict the transitions over longer time periods. In Figure \ref{fig:Dwell_drift_diff}, we present the effective drift and diffusion determined by the generative model, which are closely aligned with the true drift and diffusion. The relative errors for diffusion functions is on the order of $10^{-3}$. Figure \ref{fig:Dwell_temporal_pdf} illustrates the one-step conditional distribution $p_{X_{t+\Delta t}|X_t}(x_{t+\Delta t}|x_{t}=1.5)$ for both generative model and exact SDE. The predicted conditional distribution accurately approximates the exact one.

\subsection{SDEs with non-Gaussian noise} This section considers the learning of SDEs driven by non-Gaussian stochastic process.
\subsubsection{Noise with Exponential Distribution} 
The SDE with exponentially distributed noise is defined by
\begin{equation}\label{eq: NoiseEP_SDE}
    dX_t = \mu X_t dt +\sigma \sqrt{dt} \eta_{t}, \qquad \eta_{t}\sim \exp(1),
\end{equation}
where $\eta_{t}$ has an exponential PDF $f_{\eta}(x)=e^{-x}$, $x\geq 0$, and $\mu=-2.0$, $\sigma = 0.1$. The observation dataset $\mathcal{D}_{\rm obs}$ consists of $M=15,000K$ data pairs from $H=150,000$ trajectories, obtained by solving the SDE in Eq.~\eqref{eq: NoiseEP_SDE} with initial values uniformly sampled from $\mathcal{U}(0.2,0.9)$ up to $T=1.0$. We randomly choose $60,000$ initial states from sample $\mathcal{D}_{\rm obs}$ to generate the corresponding labeled training data. After training the generative model $G_\theta$, we simulate 500,000 prediction trajectories up to $T=5.0$.

\begin{figure}[h!]
    \centering
    \includegraphics[width=0.9\textwidth]{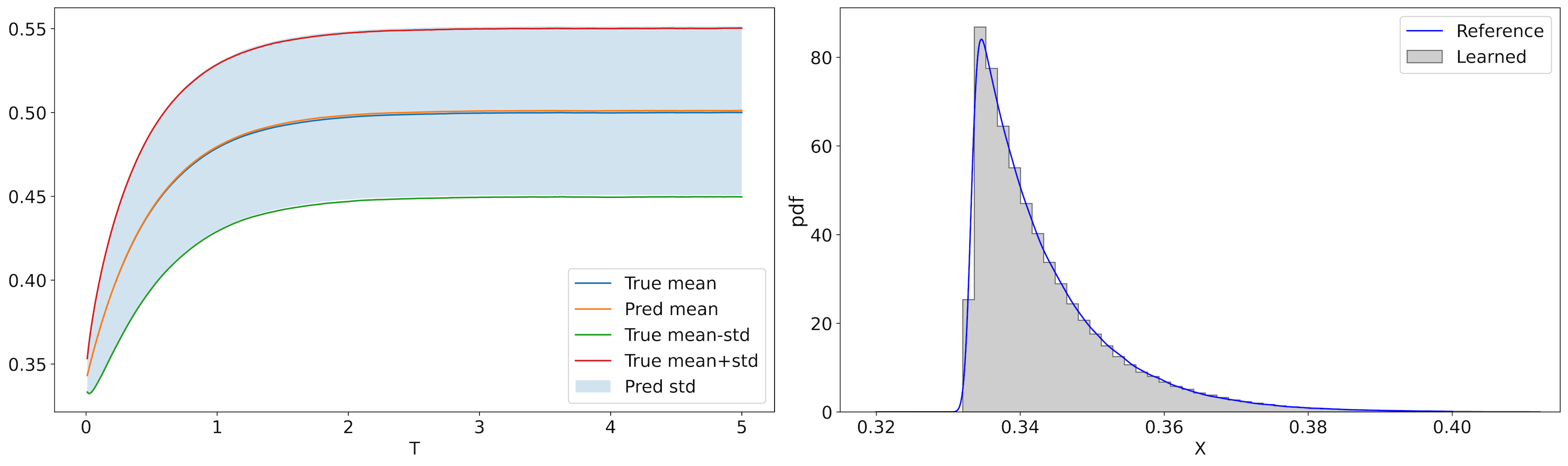}\vspace{-0.15in}
    \caption{SDE with exponential distributed noise: Left: comparison of mean and standard deviation of solutions with $x_{0}=0.34$; Right: comparison of conditional distribution $p_{X_{t+\Delta t}|X_t}(x_{t+\Delta t}|x_{t}=0.34)$.}
    \label{fig:NoiseEP_combined}
\end{figure}

\begin{figure}[!ht]
    \centering
    \includegraphics[width=0.8\textwidth]{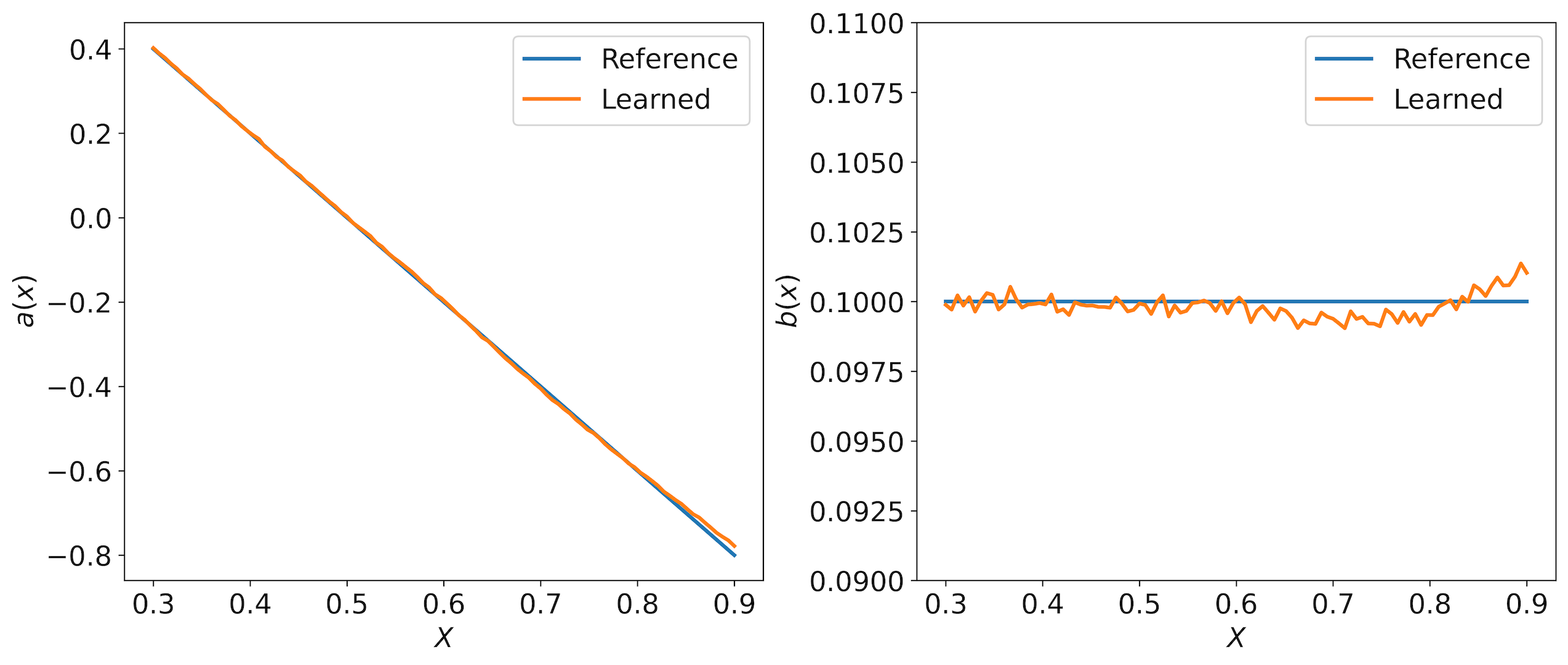} \vspace{-0.1in}
    \caption{SDE with exponential distributed noise: comparison of effective drift and diffusion functions obtained by the simulated trajectories using the generative model and the exact SDE. Left: drift $a(x)=\mu x+\sigma/\sqrt{\Delta t}$; Right: diffusion $b(x)=\sigma$.}
    \label{fig:NoiseEP_drift_diff}
\end{figure}

The mean and standard deviation of the predicted solutions for $X_{0}=0.34$ by the generative model, compared with those of the exact solutions, are shown on the left side of Figure \ref{fig:NoiseEP_combined}. The right side of Figure \ref{fig:NoiseEP_combined} illustrates the one-step conditional distribution $p_{X_{t+\Delta t}|X_t}(x_{t+\Delta t}|x_{t}=0.34)$ for both the generative model and the exact SDE. The predicted conditional distribution accurately approximates the exact one. Good agreements between the mean and standard deviation of the predicted solutions and that of the exact solution can be observed for time up to $T=5.0$ despite training dataset being limited to $T=1.0$. 
The exact effective drift and diffusion functions are
\begin{equation}\label{eq: NoiseEP_driff_diff_exact}
\begin{aligned}
       a(x) &= \mathbb{E} \left( \frac{X_{t+\Delta t}-X_{t}}{\Delta t} \middle| X_{t}=x\right)  = \mu x+ \frac{\sigma}{\sqrt{\Delta t}}, \\  
       b(x) &= \text{Std} \left( \frac{X_{t+\Delta t}-X_{t}}{\sqrt{\Delta t}}  \middle| X_{t}=x\right)  = \sigma.
\end{aligned}
\end{equation}
The corresponding effective drift and diffusion by the generative model $G_{\theta}(x,z)$ can be given by
\begin{equation}\label{eq: NoiseEP_driff_diff_learned}
    \hat{a}(x)=  \mathbb{E}_{z} \left( \frac{G_{\theta}(x,z)}{\Delta t} \right)   , \qquad  \hat{b}(x) = \text{Std}_{z} \left( \frac{G_{\theta}(x,z)}{\sqrt{\Delta t}}\right).
\end{equation}
In Figure \eqref{fig:NoiseEP_drift_diff}, we present the effective drift and diffusion determined by the generative model, which are closely aligned with the true drift and diffusion. The relative errors for diffusion functions is on the order of $10^{-2}$. 

\subsubsection{Noise with Lognormal Distribution} 
The SDE with a Lognormal distributed noise is defined by
\begin{equation}\label{eq: NoiseLN_SDE}
    d \log X_t = (\log m -\theta \log X_t) dt +\sigma dW_{t},
\end{equation}
where $m=1/\sqrt{e}$, and $\theta=1.0$, $\sigma = 0.3$. The observation dataset $\mathcal{D}_{\rm obs}$ consists of $M=20,000$K data pairs from $H=200,000$ trajectories, obtained by using the Euler-Maruyama method with the following scheme
\begin{equation}\label{eq: NoiseLN_scheme}
    X_{t+\Delta t} = m^{\Delta t} X_{t}^{1-\theta\Delta t} \eta_{t}^{\sigma\sqrt{\Delta t}}, \qquad \eta_{t} \sim \text{Lognormal}(0,1).
\end{equation}
starting from initial values that follows $\mathcal{U}(0.1,2.0)$ up to $T=1.0$. We randomly choose $60,000$ initial states from sample $\mathcal{D}_s$ to generate the corresponding labeled training data. After training the generative model $G_\theta$, we simulate 500,000 prediction trajectories up to $T=5.0$.

\begin{figure}[!ht]
    \centering
    \includegraphics[width=0.9\textwidth]{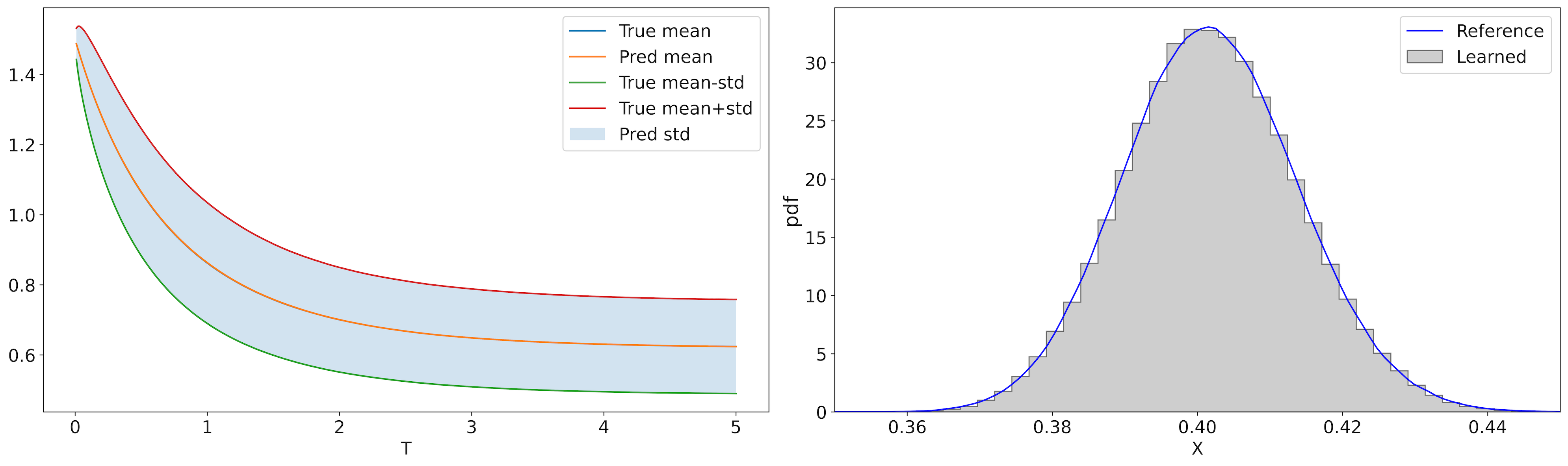}\vspace{-0.15in}
    \caption{SDE with exponential distributed noise: Left: comparison of mean and standard deviation of solutions with $x_{0}=0.4$; Right: comparison of conditional distribution $p_{X_{t+\Delta t}|X_t}(x_{t+\Delta t}|x_{t}=0.4)$.}
    \label{fig:NoiseLN_combined}
\end{figure}

The mean and standard deviation of the predicted solutions for $X_{0}=0.4$ by the generative model, compared with those of the exact solutions are displayed on the left side of Figure \ref{fig:NoiseLN_combined}. The right side of Figure \ref{fig:NoiseLN_combined} illustrates the one-step conditional distribution $p_{X_{t+\Delta t}|X_t}(x_{t+\Delta t}|x_{t}=0.4)$ for both generative model and exact SDE. The predicted conditional distribution accurately approximates the exact one. Good agreements between the mean and standard deviation of the predicted solutions and that of the exact solution can be observed for time up to $T=5.0$ despite training dataset being limited to $T=1.0$. 

\begin{figure}[!ht]
    \centering
    \includegraphics[width=0.8\textwidth]{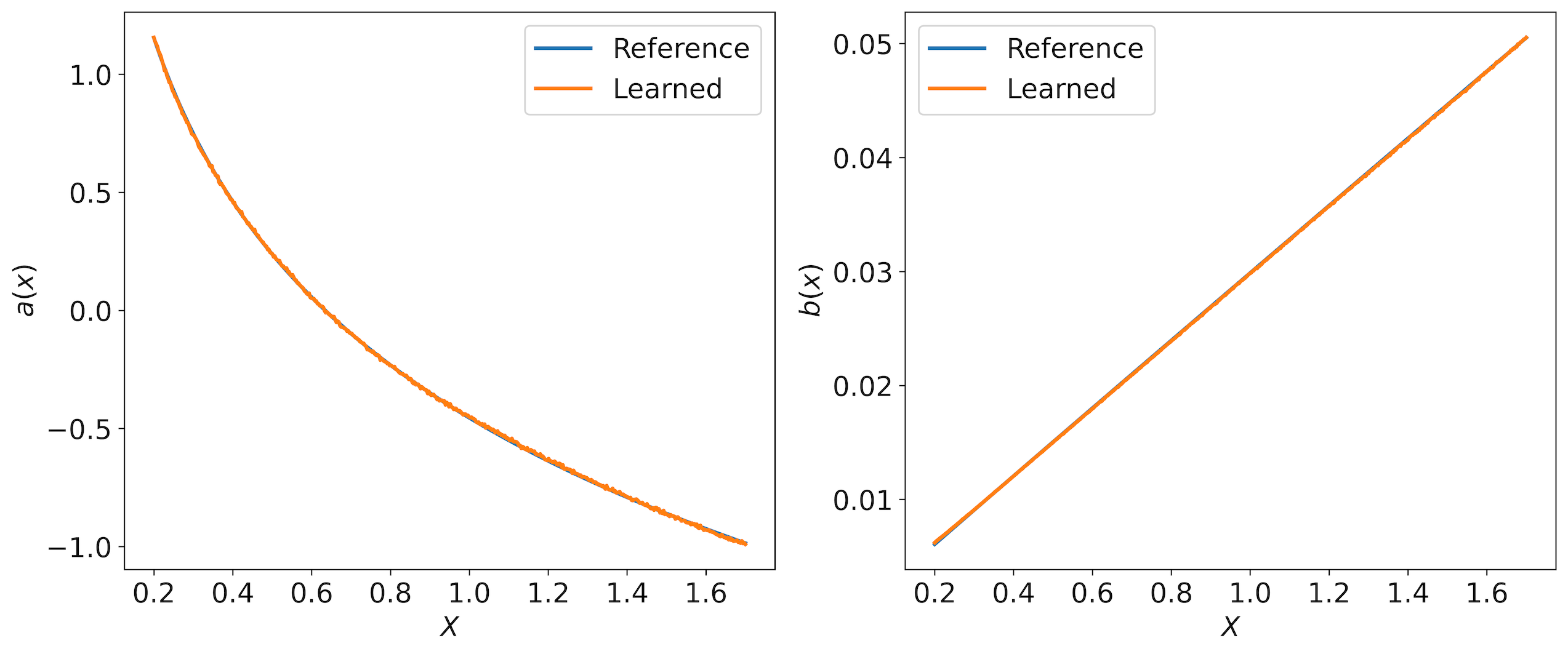} \vspace{-0.1in}
    \caption{SDE with exponential distributed noise: comparison of effective drift and diffusion functions obtained by the simulated trajectories using the generative model and the exact SDE. Left: drift $a(x)$; Right: diffusion $b(x)$. (See Eq.\eqref{eq: NoiseLN_driff_diff_exact} for the exact drift and diffusion functions and Eq.\eqref{eq: NoiseLN_driff_diff_learned} for the learned ones.)}
    \label{fig:NoiseLN_drift_diff}
\end{figure}

The SDE is rewritten in the form of the classical SDE and its effective drift and diffusion functions are
\begin{equation}\label{eq: NoiseLN_driff_diff_exact}
\begin{aligned}
       a(x) &=\ln \left[     \left( \mathbb{E} \left( \frac{X_{t+\Delta t}}{X_{t}} \middle| X_{t}=x\right)  \right)  ^{1/\Delta t}          \right ] = \ln\left( m x^{-\theta} \right) + \frac{\sigma^2}{2}, \\  
       b(x) &= \text{Std} \left(X_{t+\Delta t}|X_{t}=x\right) = \sqrt{ e^{\sigma^2\Delta t}-1} \left(m e^{\sigma^2/2}\right)^{\Delta t} x^{1-\theta \Delta t}.
\end{aligned}
\end{equation}
The corresponding effective drift and diffusion by the generative model $G_{\theta}(x,z)$ can be given by
\begin{equation}\label{eq: NoiseLN_driff_diff_learned}
    \hat{a}(x) =\ln \left[     \left( \mathbb{E}_{z} \left( \frac{G_{\theta}(x,z)+x}{x} \right)  \right)  ^{1/\Delta t}          \right ] , \qquad  \hat{b}(x) = \text{Std}_{z} \left(G_{\theta}(x,z)\right).
\end{equation}
In Figure \ref{fig:NoiseLN_drift_diff}, we present the effective drift and diffusion determined by the generative model, which are closely aligned with the true drift and diffusion. The relative errors for diffusion functions is on the order of $10^{-2}$. 

\subsection{Multi-Dimensional Examples} In this section, we consider the learning of multi-dimensional SDE systems.

\subsubsection{Two-dimensional Ornstein-Uhlenbeck process} The first example is the two-dimensional OU process
\begin{equation} \label{eq: 2DOU_SDE}
    d{ {X}}_{t}= {  {B}} { {X}}_{t} dt + {  {\Sigma}}~ d {  {W}}_{t},
\end{equation}
where ${ {X}}_t = (x_1,x_2)\in \mathbb{R}^2$ are the state variables, ${  {B}}$ and ${  {\Sigma}}$ are the following $2\times2$ matrices,
\begin{equation}\label{eq: 2DOU_coefficients}
    {  {B}} =\begin{pmatrix}
        -1 & -0.5\\
        -1 & -1
    \end{pmatrix}, \qquad 
       {  {\Sigma}} =\begin{pmatrix}
        1 & 0\\
        0 & 0.5
    \end{pmatrix}.
\end{equation}

The observation dataset $\mathcal{D}_s$ consists of $M=35,000$K data pairs from $H=350,000$ trajectories, obtained by solving the SDE in Eq.~\eqref{eq: 2DOU_SDE} under Eq.~\eqref{eq: 2DOU_coefficients} with initial values uniformly sampled from $(-4,4)\times(-3,3)$ up to $T=1.0$. We randomly choose $120,000$ initial states from sample $\mathcal{D}_{\rm obs}$ to generate the corresponding labeled training data. After training the generative model $G_\theta$, we simulate 500,000 prediction trajectories up to $T=5.0$.

\begin{figure}[!ht]
    \centering
    \includegraphics[width=0.9\textwidth]{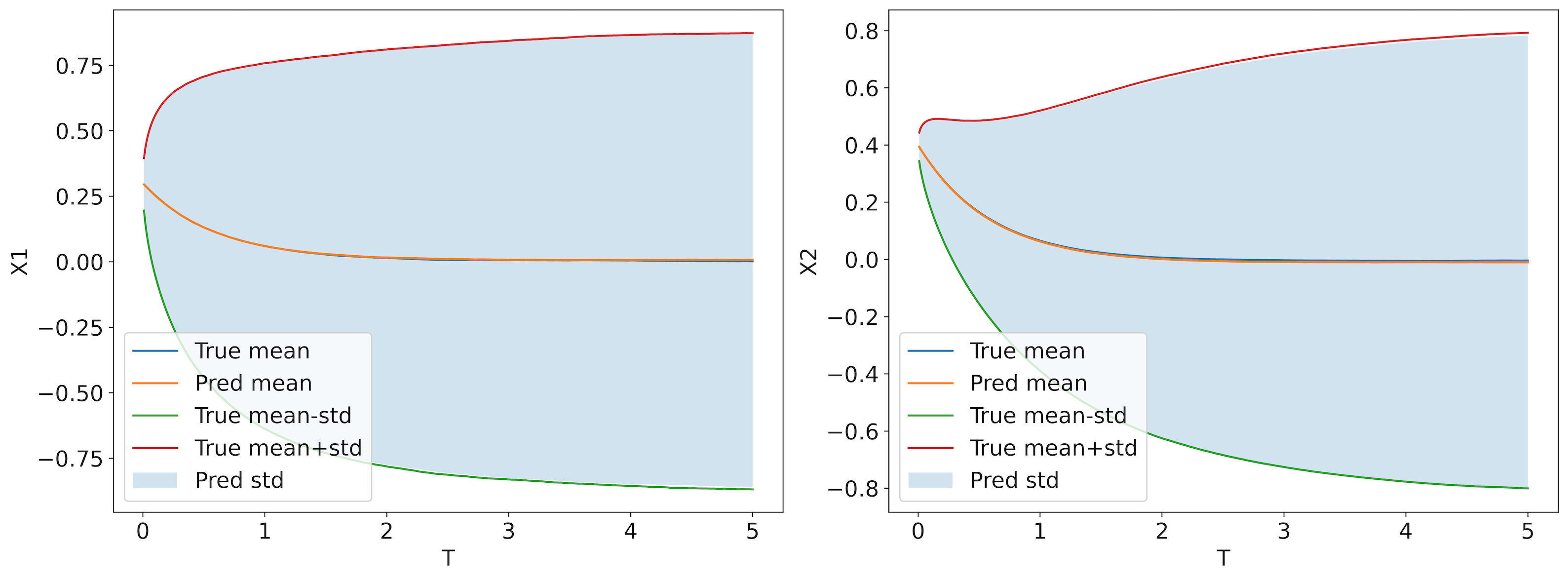} \vspace{-0.15in}
    \caption{2D OU: comparison of mean and standard deviation of solutions with ${ {X}}_{0}=(0.3,0.4)$. }
    \label{fig:2DOU_mean_std}
\end{figure}

\begin{figure}[!ht]
    \centering
    \includegraphics[width=0.9\textwidth]{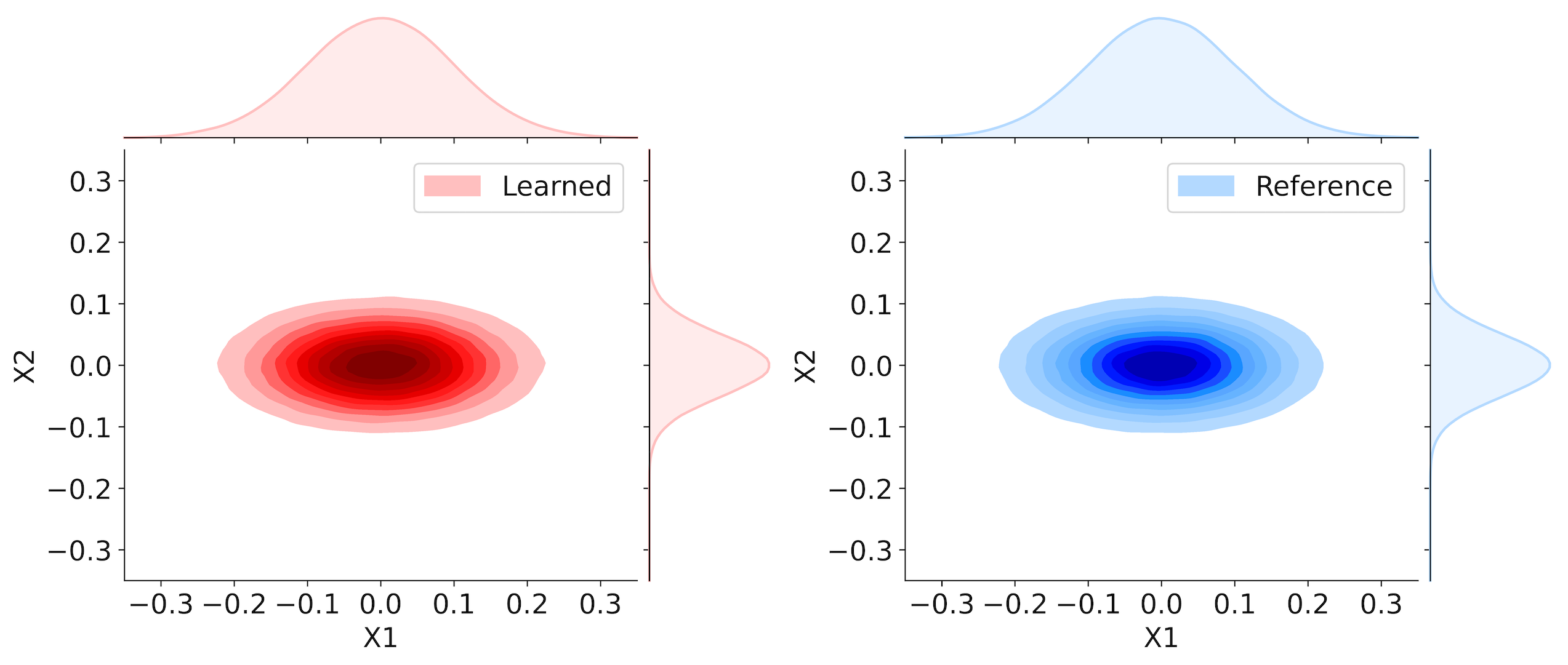}\vspace{-0.15in}
    \caption{2D OU: Left: comparison of conditional PDF $p_{X_{t+\Delta t}|X_t}(x_{t+\Delta t}|x_{t}=(0,0))$ determined by the generative model $G_\theta$ and the exact flow map $F_{\Delta t}$. Left: learned; Right: reference.}
    \label{fig:2DOU_contour_plot}
\end{figure}

The mean and standard deviation of the predicted solutions for ${{X}}_{0}=(0.3,0.4)$ by the generative model, compared with those of the exact solutions are displayed in Figure \ref{fig:2DOU_mean_std}. Figure \ref{fig:2DOU_contour_plot} illustrates the one-step conditional probability 
distribution $p_{X_{t+\Delta t}|X_t}(x_{t+\Delta t}|x_{t}=(0,0))$ for both generative model and exact SDE. The predicted conditional distribution accurately approximates the exact one. Good agreements between the mean and standard deviation of the predicted solutions and that of the exact solution can be observed for time up to $T=5.0$ despite training dataset being limited to $T=1.0$. 

\subsubsection{Two-dimensional stochastic oscillator} The second example is the stochastic oscillator with 
\begin{equation}\label{eq: 2DSO_coefficients}
    {  {B}} =\begin{pmatrix}
       0 & 1\\
        -1 & 0
    \end{pmatrix}, \qquad 
       {  {\Sigma}} =\begin{pmatrix}
        0 & 0\\
        0 & 0.1
    \end{pmatrix}
\end{equation}
in Eq.~\eqref{eq: 2DOU_SDE}.
The observation dataset $\mathcal{D}_{\rm obs}$ consists of $M=300,000$K data pairs from $H=3,000,000$ trajectories, obtained by solving the SDE in Eq.~\eqref{eq: 2DOU_SDE} under Eq.~\eqref{eq: 2DSO_coefficients} with initial values uniformly sampled from $(-1.5,1.5)\times(-1.5,1.5)$ up to $T=1.0$. We randomly choose $50,000$ initial states from sample $\mathcal{D}_{\rm obs}$ to generate the corresponding labeled training data. After training the generative model $G_\theta$, we simulate 500,000 prediction trajectories up to $T=6.5$.

The mean and standard deviation of the predicted solutions for ${ {X}}_{0}=(0.3,0.4)$ by the generative model, compared with those of the exact solutions are displayed in Figure \ref{fig:2DOU_mean_std}. Figure \ref{fig:2DOU_contour_plot} illustrates the one-step conditional probability distribution $p_{X_{t+\Delta t}|X_t}(x_{t+\Delta t}|x_{t}=(-0.5,-0.5))$ for both generative model and exact SDE. The predicted conditional distribution accurately approximates the exact one. Good agreements between the mean and standard deviation of the predicted solutions and that of the exact solution can be observed for time up to $T=6.5$ despite the training dataset being limited to $T=1.0$. 
\begin{figure}[!ht]
    \centering
    \includegraphics[width=0.9\textwidth]{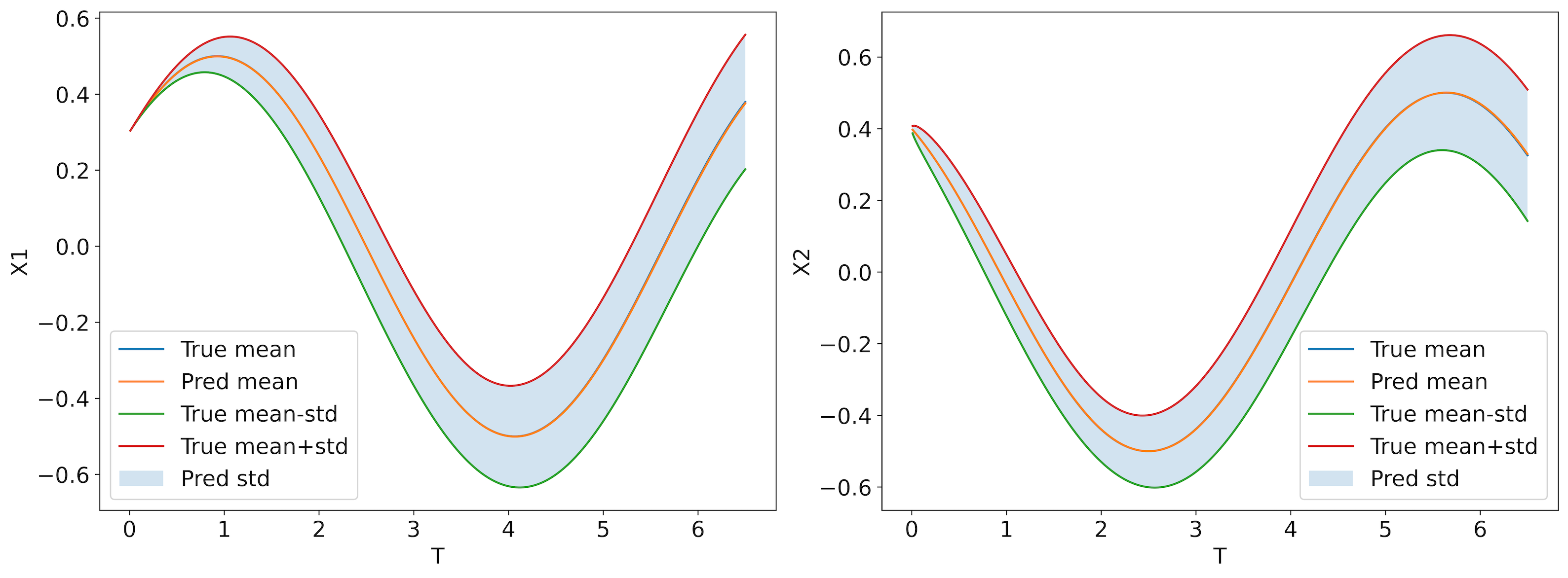} \vspace{-0.15in}
    \caption{2D stochastic oscillator: comparison of mean and standard deviation of solutions at ${ {X}}_{0}=(0.3,0.4)$. Left: $x_{1}$; Right: $x_{2}$.}
    \label{fig:2DSO_mean_std}
\end{figure}
\begin{figure}[!ht]
    \centering
    \includegraphics[width=0.9\textwidth]{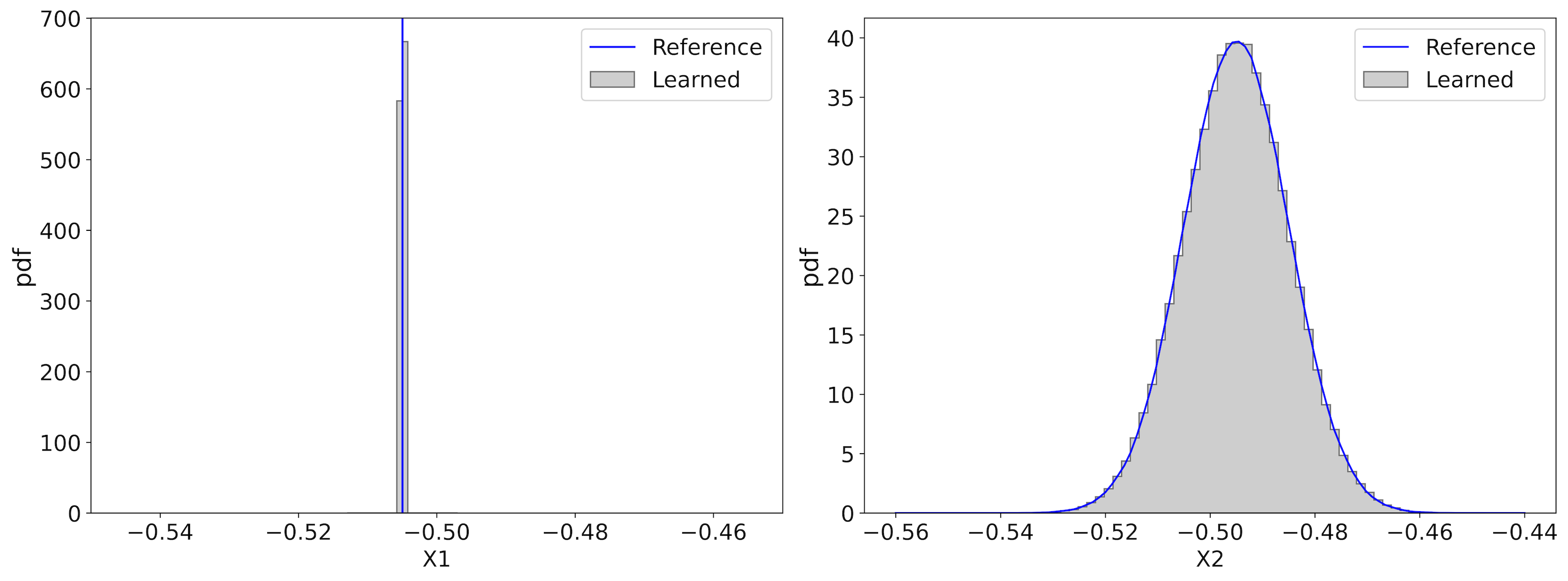}\vspace{-0.15in}
    \caption{2D stochastic oscillator: comparison of marginal probability distribution at ${ {X}}_{0}=(-0.5,-0.5)$. Left: $x_{1}$; Right: $x_{2}$.}
    \label{fig:2DSO_marginal}
\end{figure}

\subsubsection{Five-dimensional Ornstein-Uhlenbeck process} This example is the 5-dimensional OU process, driven by Wiener processes of varying dimensions
\begin{equation} \label{eq: 5DOU_SDE}
    d{ {X}}_{t}= {  {B}} { {X}}_{t} dt + {  {\Sigma}}~ d {  {W}}_{t},
\end{equation}
where ${ {X}}_t = (x_1,\cdots, x_5)\in \mathbb{R}^5$ are the state variables, ${  {B}}$ and ${  {\Sigma}}$ are the following $5\times5$ matrices. The 
\begin{equation}\label{eq: 5DOU_B}
    {  {B}} =\begin{pmatrix}
0.2 & 1.0 & 0.2 & 0.4 & 0.2 \\
-1.0 & 0.0 & 0.2 & 0.8 & -1.0 \\
0.2 & 0.2 & -0.8 & -1.2 & 0.2 \\
-0.6 & 0.0 & 1.2 & -0.2 & 0.6 \\
0.2 & 0.2 & 0.6 & 0.4 & 0.0 \\
    \end{pmatrix}.
\end{equation}

We consider the following $5$ different scenarios for ${  {\Sigma}}$, which has rank varying from $1$ to $5$. The matrices for ${  {\Sigma}}$ are 
\begin{equation*}
     {  {\Sigma_1}} =\text{diag}(0,0,1,0,0), 
     \end{equation*}
     \begin{equation*}
        {  {\Sigma_2}} =\text{diag}(0,0.8,0,0,-0.8),~ {  {\Sigma_3}} =\begin{pmatrix}
       0.8 & 0.2 & 0.0 & 0.0 & 0.0 \\
-0.4& 0.6 & 0.0 & 0.0 & 0.0 \\
0.0 & 0.0 & 0.0 & 0.0 & 0.0 \\
0.0 & 0.0 & 0.0 & 0.7 & 0.0 \\
0.0 & 0.0 & 0.0 & 0.0 & 0.0 \\
    \end{pmatrix},      
    \end{equation*}
     \begin{equation*}
    {  {\Sigma_4}} =\begin{pmatrix}
      0.7 & 0.0 & -0.4 & 0.0 & 0.0 \\
0.0 & 0.0 & 0.0 & 0.0 & 0.0 \\
0.1 & 0.0 & 0.6 & 0.2 & -0.1 \\
0.0 & 0.0 & 0.1 & -0.6 & 0.2 \\
0.0 & 0.0 & 0.0 & 0.3 & 0.8 \\
    \end{pmatrix}, ~
    {  {\Sigma_5}} = \begin{pmatrix}
0.8 & 0.2 & 0.1 & -0.3 & 0.1 \\
-0.3 & 0.6 & 0.1 & 0.0 & -0.1 \\
0.2 & -0.1 & 0.9 & 0.1 & 0.2 \\
0.1 & 0.1 & -0.2 & 0.7 & 0.0 \\
-0.1 & 0.1 & 0.1 & -0.1 & 0.5 \\
\end{pmatrix}. \\
\end{equation*}
The observation dataset $\mathcal{D}_{\rm obs}$ consists of $M=300,000$K data pairs from $H=3,000,000$ trajectories, obtained by solving the SDE in Eq.~\eqref{eq: 5DOU_SDE} with initial values uniformly sampled from the hypercube $(-1.0,1.0)^5$ up to $T=1.0$. We randomly choose $50,000$ initial states from sample $\mathcal{D}_{\rm obs}$ to generate the corresponding labeled training data. After training the generative model $G_\theta$, we simulate 500,000 prediction trajectories up to $T=5.0$.

\begin{figure}[!ht]
    \centering
    \includegraphics[width=\textwidth]{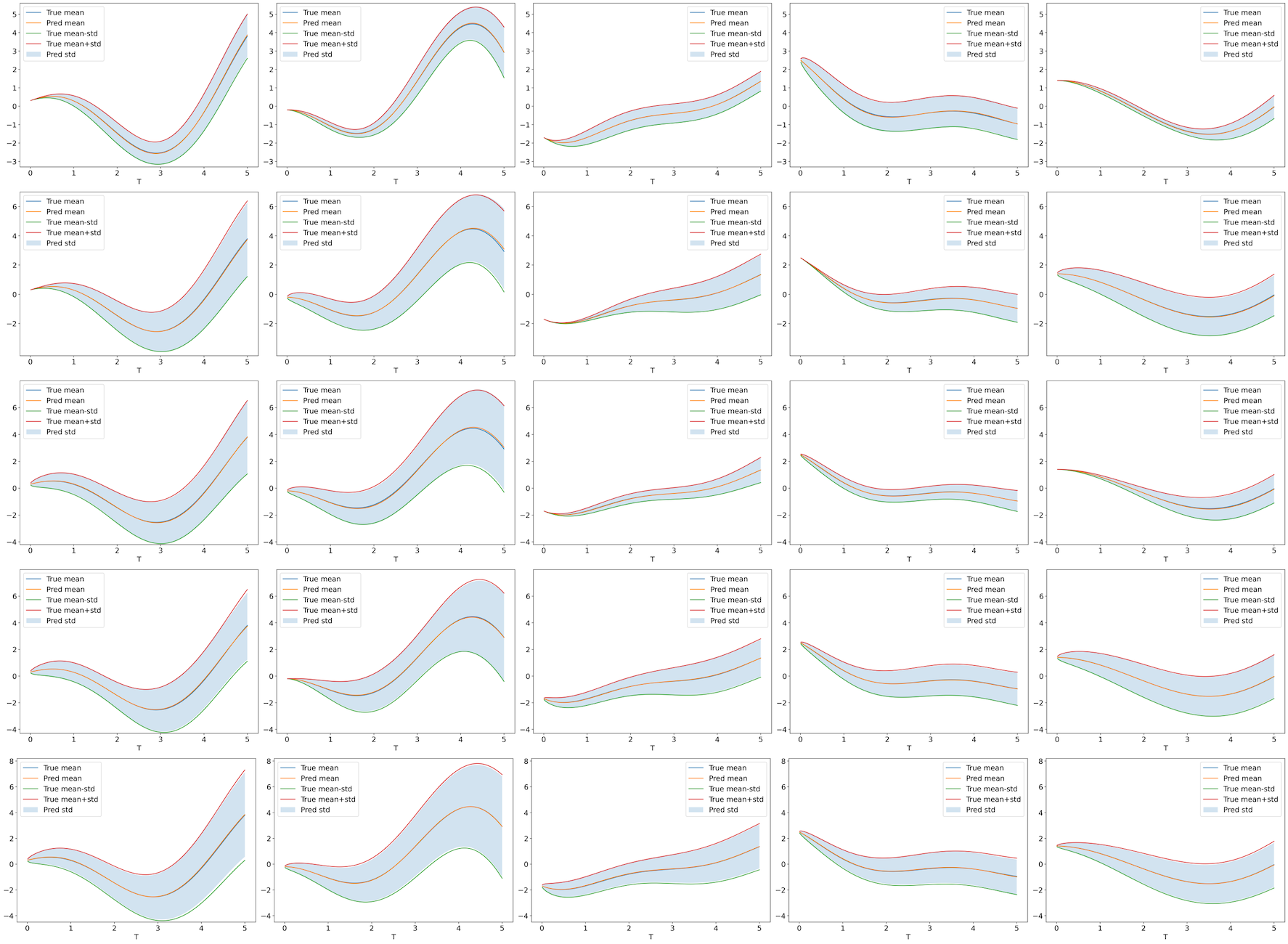} \vspace{-0.15in}
    \caption{5D OU process: comparison of mean and standard deviation of solutions at ${ {X}}_{0}=(0.3, -0.2,$ $ -0.7, 0.5, 0.6)$. Left to right: $x_{1}, x_{2},\cdots, x_{5}$.
    Top to bottom: $\Sigma_{1}, \Sigma_{2},\cdots, \Sigma_{5}$.}
    \label{fig:5D_mean_std}
\end{figure}

\begin{figure}[!ht]
    \centering
    \includegraphics[width=\textwidth]{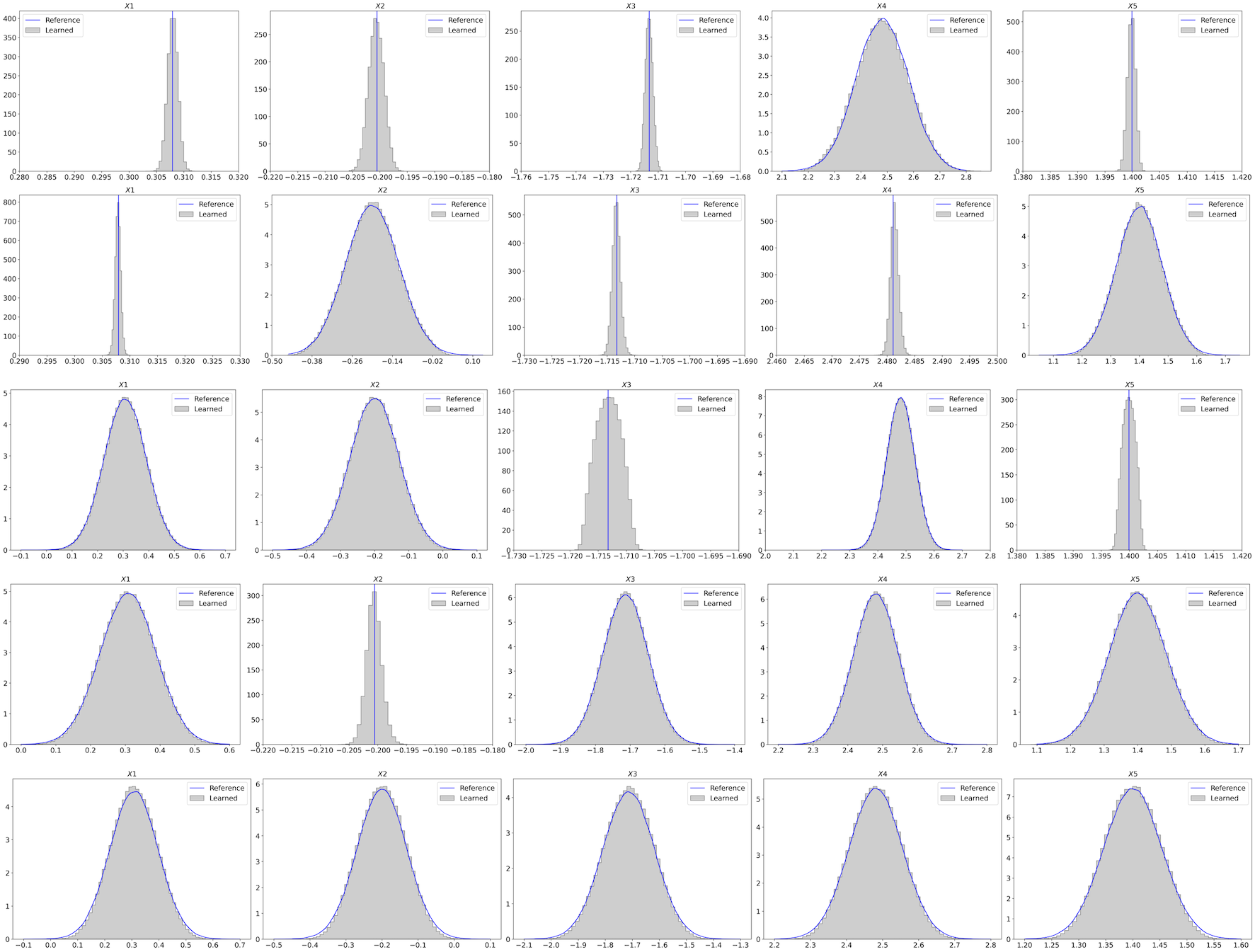}\vspace{-0.15in}
    \vspace{0.3cm}
    \caption{5D OU process: comparison of marginal probability distribution at ${ {X}}_{0}=(0.3, -0.2,$ $ -0.7, 0.5, 0.6)$. Left to right: $x_{1}, x_{2},\cdots, x_{5}$.
    Top to bottom: $\Sigma_{1}, \Sigma_{2},\cdots, \Sigma_{5}$.}
    \label{fig:5D_marginal}
\end{figure}

The mean and standard deviation of the predicted solutions for initial value ${ {X}}_{0}=(0.3, -0.2,$ $ -0.7, 0.5, 0.6)$ by the generative model, compared with those of the exact solutions are displayed in Figure \ref{fig:5D_mean_std}. Figure \ref{fig:5D_marginal} illustrates the one-step conditional distribution $p_{X_{t+\Delta t}|X_t}(x_{t+\Delta t}|x_{t}=X_0)$ for both the generative model and the exact SDE. The predicted conditional distribution accurately approximates the exact one. Good agreements between the mean and standard deviation of the predicted solutions and that of the exact solution can be observed for time up to $T=6.5$ despite training dataset being limited to $T=1.0$. 

\subsection{Comparison with baseline methods}
In this section, we compare the numerical performance of the proposed method with baseline: the GANs model proposed in \cite{CHEN2024112984}. The following metrics are considered for comparison of model accuracy:
\begin{itemize}[leftmargin=15pt]\itemsep0cm
    \item (Relative) error of (1D) effective drift function: $E_a = \|a(x)-\hat{a}(x)\|/\|a(x)\|$;
    \item (Relative) error of (1D) effective diffusion function: $E_b = \|b(x)-\hat{b}(x)\|/\|b(x)\|$;
    \item Error of the predictive mean at termination time: $E_T^m = \|\mathbb{E}(\mathbf{x}_T)-\mathbb{E}(\hat{\mathbf{x}}_T)\|$;
    \item Error of the predictive STD at termination time: $E_T^s = \|STD(\mathbf{x}_T)-STD(\hat{\mathbf{x}}_T)\|$.
\end{itemize}

The comparison results are summarized in Table \ref{driftdiffusion} and \ref{meanstdest}. In the tables, we highlight the better results. We observed that, for all 4 metrics, our method outperforms the baseline in most examples. 
It is also worth noting that our method is more efficient than GANs for training. The GANs modeling of sFML suffers from many training difficulties, which are inherited from Vanilla GANs, such as the requirement of a huge number of epochs (estimated $\mathcal{O}(10^5)$), lack of reliable metrics indicating when to terminate, etc. In contrast, our method, the time-consuming step is the generation of labeled data, which only takes a few minutes. Moreover, this single component can be further accelerated by paralleling sampling on the phase space of the learned SDE.
\begin{table}[h!]
\caption{Comparison of effective drift and diffusion functions estimation for GANs and our method.}
\label{driftdiffusion}
\resizebox{\textwidth}{!}{
\begin{tabular}{lllll}
\toprule
\textbf{Example}          & \textbf{$E_a$ of GANs} & \textbf{$E_a$ of our method} & \textbf{$E_b$ of GANs} & \textbf{$E_b$ of our method} \\ \hline
OU process                & 3.7615E-02             &  \textbf{2.0264E-02}                     & 3.4912E-03             &    \textbf{2.5218E-03}                    \\
Geometric Brownian Motion & 3.8360E-02             &     \textbf{1.3084E-02}                    & 3.0818E-02             &  \textbf{2.1089E-02}                    \\
Exp diffusion SDE         & 2.9407E-02             &    \textbf{2.0723E-03}                    & 1.4196E-02             &     \textbf{9.5213E-03}                   \\
Trigonometric SDE         & 3.8318E-02             &      \textbf{2.4540E-02}                 & 3.8146E-02             &   \textbf{3.4354E-03}                     \\
Double Well Potential     & 5.6902E-02             &   \textbf{1.2980E-02}                     & 1.8266E-02             &   \textbf{2.0697E-03}                     \\
Exp. Noise SDE            & 2.4217E-02             & \textbf{1.6449E-02}                       & 3.5840E-02             &      \textbf{4.9811E-03}                  \\
Lognormal Noise SDE       & 6.7650E-02             &      \textbf{9.5213E-03}                  & 1.9651E-02             &            \textbf{2.0723E-03}            \\ \bottomrule
\end{tabular}
}
\end{table}
\begin{table}[h!]
\caption{Comparison of SDE end-time mean and STD estimation for GANs and our method.}
\label{meanstdest}
\resizebox{\textwidth}{!}{
\begin{tabular}{llllll}
\toprule
\textbf{Example}          & $\boldsymbol{T}$ & \textbf{$E_T^m$ of GANs} & \textbf{$E_T^m$ of our method} & \textbf{$E_T^s$ of GANs} & \textbf{$E_T^s$ of our method} \\ \hline
OU process                & 4  & \textbf{1.3538E-03}             &   3.6823E-03                     & 7.3188E-04             &  \textbf{5.4972E-04}                      \\
Geometric Brownian Motion & 1  & 1.0389E-01             &    \textbf{3.1599E-02}                   & 2.3274E-01             &      \textbf{8.9173E-02}                   \\
Exp diffusion SDE         & 10  & 4.0665E-03             &   \textbf{1.8914E-04}                     & \textbf{4.3357E-04}             &            7.4537E-04            \\
Trigonometric SDE         & 10  & 6.6718E-04             &   \textbf{2.8130E-04}                      & 1.7917E-03             &    \textbf{2.2179E-04}                    \\
Double Well Potential     &  300  & 1.2037E-01            &     \textbf{1.7988E-02}                   & 2.5084E-02             &     \textbf{2.2183E-03}                   \\
Exp. Noise SDE            &  5  & 1.8862E-03             &   \textbf{1.1758E-03}                     & 4.4275E-04             &   \textbf{2.6745E-05}                     \\
Lognormal Noise SDE       &  5  & 9.2052E-03             &      \textbf{1.8914E-04}                  & 3.0957E-03             &               \textbf{7.4537E-04}         \\
2D OU process             &  5  & 3.8093E-02             &      \textbf{8.8535E-03}                  & 2.4988E-02             &    \textbf{5.1506E-03}                    \\
Stochastic Oscillator     &  6.5  & 1.7828E-02             &    \textbf{1.8628E-03}                    & 1.1120E-02             &      \textbf{3.7311E-04}                  \\ \bottomrule
\end{tabular}
}
\end{table}

\section{Conclusion}\label{sec:con}
In this paper, we have introduced a novel training-free conditional diffusion model for learning the flow map of stochastic dynamical systems. Our approach offers a promising alternative to traditional methods for solving SDEs, addressing key challenges in computational efficiency and accuracy. The proposed method leverages the power of score-based diffusion models to generate samples from complex distributions without the need for extensive training data or computationally intensive solving procedures. By introducing a conditional framework, we have enabled our model to capture the dynamics of SDEs across a wide range of initial conditions and system parameters. Our numerical experiments, spanning various types of SDEs including linear, nonlinear, and multi-dimensional systems, demonstrate the effectiveness and versatility of our approach. The model shows remarkable accuracy in predicting both short-term and long-term behaviors of stochastic systems, often outperforming baseline methods such as GANs in terms of drift and diffusion function estimation, as well as in predicting mean and standard deviation at termination times. 

While our proposed method has shown promising results, there are several avenues for future research that could further enhance its capabilities and applications. First, extending the model to handle more complex stochastic processes, such as jump diffusions or fractional Brownian motion, could broaden its applicability in fields like finance and environmental science. Second, investigating the theoretical properties of the proposed score estimation technique, including convergence rates and error bounds, would provide a stronger mathematical foundation for the method. Third, exploring the integration of this approach with other machine learning techniques, such as transfer learning or meta-learning, could potentially improve its performance on new, unseen stochastic systems. Fourth, applying this method to real-world problems in areas such as climate modeling, epidemiology, or quantum systems could demonstrate its practical value and potentially lead to new insights in these fields. Finally, optimizing the computational efficiency of the algorithm, particularly for high-dimensional systems, remains an important area for future work. This could involve developing advanced parallelization strategies or leveraging specialized hardware accelerators to further reduce computation time.

%
\end{document}